\documentclass[10pt,journal]{IEEEtran}

\usepackage{amsmath,amssymb}
\usepackage{graphicx}
\usepackage{booktabs}
\usepackage{multirow}
\usepackage{array}
\usepackage{siunitx}
\usepackage{cite}
\usepackage{url}
\usepackage[hidelinks]{hyperref}
\usepackage[caption=false,font=footnotesize]{subfig}
\usepackage{balance}

\graphicspath{{figures/}}
\title{Hybrid Roller-Jamming Gripper for Object Acquisition and Retention Under Pose Uncertainty}

\author{Yijie Ren, Guillaume Gourmelen, and Hiroyasu Iwata%
\thanks{Yijie Ren, Guillaume Gourmelen, and Hiroyasu Iwata are with the Department of Modern Mechanical Engineering, Graduate School of Creative Science and Engineering, Waseda University, Tokyo, Japan.}%
\thanks{Corresponding author: Yijie Ren (e-mail : yijie.ren@ruri.waseda.jp), Guillaume Gourmelen (email : g.gourmelen@kurenai.waseda.jp)}%
}

\begin{document}

\maketitle

% =========================================================
% ABSTRACT
% =========================================================
\begin{abstract}
In household manipulation, pose uncertainty often results in off-centre or partial initial contact, making reliable object acquisition difficult. Roller-based grippers can actively draw objects inward but often provide limited post-capture stability, whereas granular-jamming grippers require sufficient contact before jamming to achieve strong retention. This paper presents a hybrid roller-jamming gripper that integrates active object intake and post-capture retention within a single gripper. The proposed gripper uses inward roller rotation to increase contact and draw the object toward the gripper centre, followed by vacuum-induced granular jamming to stiffen the rollers and stabilise the grasp. The paper also presents a simplified geometric analysis of the gripper and a bench-level characterisation of the prototype’s force capability. The gripper prototype was mounted on a 7-DoF robotic arm and evaluated using eight test objects. Furthermore, controlled planar position and orientation offsets were applied, with each condition repeated three times. The main evaluation comprised 840 grasp trials, including 216 planar-offset trials and 624 orientation-offset trials. Overall, the gripper succeeded in 812/840 trials: 215/216 planar-offset trials and 597/624 orientation-offset trials. The ablation evaluation comprised 162 trials on three objects. The roller-only and jamming-only conditions achieved 54/81 and 24/81 successes, respectively, showing their different contributions. These results provide initial mechanism-level evidence that hybrid roller-jamming is a promising strategy for improving acquisition and retention after imperfect first contact.
\end{abstract}


%%%%%%%%%%%%%%%%%%%%%%%%%%%%%%%%%%%%%%%%%%%%%%%%%%%%%%%%%%%%%%%%%%%%%%%%%%%%%%%%

\section{Introduction}

%===============================================================================

\subsection{Motivation}
A home-assist robot may be required to pick a pill bottle from a cluttered nightstand, retrieve a balled-up sock from the floor, or move a full glass of water without spilling. Picking up dropped or misplaced objects is an important assistive capability in domestic environments. However, real homes introduce additional challenges in mobility, perception, safety, and whole-system integration compared with controlled laboratory settings \cite{de2019grasping}. Such tasks involve objects that vary widely in shape, size, stiffness, surface properties, and fragility. Industrial environments often provide known object positions and orientations. By contrast, home environments are typically cluttered, partially occluded, and visually unstructured. Consequently, object observations and pose estimates are often incomplete or noisy, leading to off-centre or partial initial contact during grasp execution. Under these conditions, reliable grasp acquisition becomes difficult, particularly when the gripper cannot effectively recover after imperfect first contact \cite{hsiao2011robust}.

%===============================================================================

\subsection{Related Work and Problem Gap}
A widely adopted strategy for robust grasping across diverse objects is to combine compliance and underactuation. This allows the gripper to passively adapt its contact geometry during contact, while avoiding the need to explicitly control every joint. Compliant hands such as the RBO Hand 2 \cite{deimel2016novel} achieve diverse grasps with low control complexity. Mode-switching underactuated hands such as the Hydra Hand \cite{chappell2023hydra} support both power and precision grasping. Hybrid designs further extend this approach by combining compliant and rigid elements. For example, passive pneumatic soft joints have been shown to improve the handling of thin, deformable objects such as paper \cite{tran2025hybrid}.

Furthermore, mechanisms that remain compliant during approach and then stiffen after contact are particularly attractive for robotic grasping because they combine gentle initial interaction with improved stability after capture. Granular jamming is a representative example of this strategy. A particle-filled membrane first conforms to the target and then stiffens under vacuum, increasing resistance to slip through frictional reinforcement and geometric interlocking \cite{amend2012positive}. Subsequent work has improved its payload, controllability, and surface compatibility through approaches such as layer jamming \cite{zeng2023high}, electroadhesive-jamming hybrids \cite{piskarev2023soft}, and phase-change emulsion jamming for micro-textured surfaces \cite{keller2024phase}. Jamming has also been integrated into two-finger grippers with jamming fingertips \cite{hou2019design} and dexterous hands such as the JamHand \cite{amend2017jamhand}. Other examples include hybrid rigid-jamming graspers for surgical instruments \cite{badilla2024hybgrip}, compact underactuated grippers with pneumatic modules \cite{kim2024development}, and tendon-driven multi-finger hands \cite{mizushima2018multi}.

Despite these advances, granular-jamming-based graspers still depend on establishing sufficient contact before stiffening can effectively secure the object. In household environments, however, clutter, occlusion, and perception error often lead to off-centre contact or approach-pose error, making this contact state difficult to establish reliably. A different line of work addresses this challenge through actively driven rollers or belts. These systems use surface motion after contact to draw the object inward, translate it within the grasp, or reorient it while maintaining contact. Examples include fingertip-roller systems for fruit handling \cite{qin2025rolling}, belted parallel-jaw grippers \cite{xie2023hand}, and roller-based hands developed for within-hand manipulation \cite{yuan2020design,yuan2020designV2,yuan2024designV3}. Such active-surface grippers demonstrate that rolling contact can convert partial engagement into a more centred, secure grasp. An omnidirectional Mecanum-wheel hand has also been proposed for in-hand manipulation without finger gaiting ~\cite{li2026dexterous}. Table~\ref{tab:gripper_comparison} compares its acquisition, retention, and actuation mechanisms with those of other representative grippers and the proposed design. Because the cited systems were evaluated using different objects and protocols, their payload values are not directly compared in Table~\ref{tab:gripper_comparison}.

Roller-based and belt-based grippers improve acquisition and post-contact repositioning. However, they do not inherently increase grasp stiffness or retention after capture, particularly for objects with slippery surfaces or complex local geometry. Conversely, granular jamming can strongly stabilise a grasp once sufficient contact has been established, but it remains sensitive to the quality of the contact state before stiffening. This motivates a hybrid strategy in which rolling is used to improve engagement during acquisition, and jamming is used to preserve that engaged state during retention. This paper focuses on mechanism-level validation of that hybrid strategy under controlled planar position and orientation offsets.

\begin{table*}[t]
\centering
\caption{Comparison with representative compliant,
jamming-based, and active-surface grippers. The systems
address different tasks and are compared primarily by their
mechanisms and actuation requirements.}
\label{tab:gripper_comparison}
\scriptsize
\setlength{\tabcolsep}{4pt}
\renewcommand{\arraystretch}{2}
\begin{tabular}{
p{2.5cm}
p{3.1cm}
p{2.8cm}
p{2.8cm}
p{4.0cm}}
\hline
\textbf{System} &
\textbf{Primary purpose} &
\textbf{Acquisition mechanism} &
\textbf{Retention mechanism} &
\textbf{Actuation architecture} \\
\hline

RBO Hand 2~\cite{deimel2016novel} &
Dexterous grasping through compliance and underactuation &
Pneumatic bending of compliant fingers and palm around the
object &
Compliant enveloping contact maintained by pneumatic pressure &
Seven pneumatic actuators grouped into four control channels \\

\hline
Two-fingered hand with jamming finger tips~\cite{hou2019design} &
Adaptive grasping with variable-stiffness fingertips &
Finger closure with passive fingertip conformity &
Vacuum-jammed fingertips &
Four servomotors and a pneumatic jamming system \\

\hline
Simple Belted Parallel-Jaw Gripper~\cite{xie2023hand} &
Grasp-to-grasp in-hand translation and reorientation &
Base parallel-jaw closure &
Force-controlled jaw clamping with belt and finger-surface
contact &
Four independent belt motors plus the base-gripper actuator \\

\hline
Roller Grasper V3~\cite{yuan2024designV3} &
Full 6-DoF in-hand manipulation &
Finger closure followed by rolling pickup &
Four-finger rolling contact, assisted by an active palm &
Four fingers with three active DoF each; 12 active DoF \\

\hline
Mecanum-wheel hand~\cite{li2026dexterous} &
Omnidirectional in-hand manipulation without finger gaiting &
Coupled four-finger opening and closing &
Force closure with spring-loaded wheel constraint forces &
Three-servo coupled reconfiguration and two wheel-drive motors \\

\hline
Proposed roller-jamming gripper &
Grasp acquisition and retention under pose offsets &
Shared finger closure followed by inward rolling intake &
Vacuum-induced granular jamming of the same rollers &
Three active DoF and pneumatic vacuum switching \\
\hline
\end{tabular}
\end{table*}

%===============================================================================

\subsection{Objectives}
\begin{figure}[thpb]
    \centering
    \includegraphics[width=0.8\linewidth]{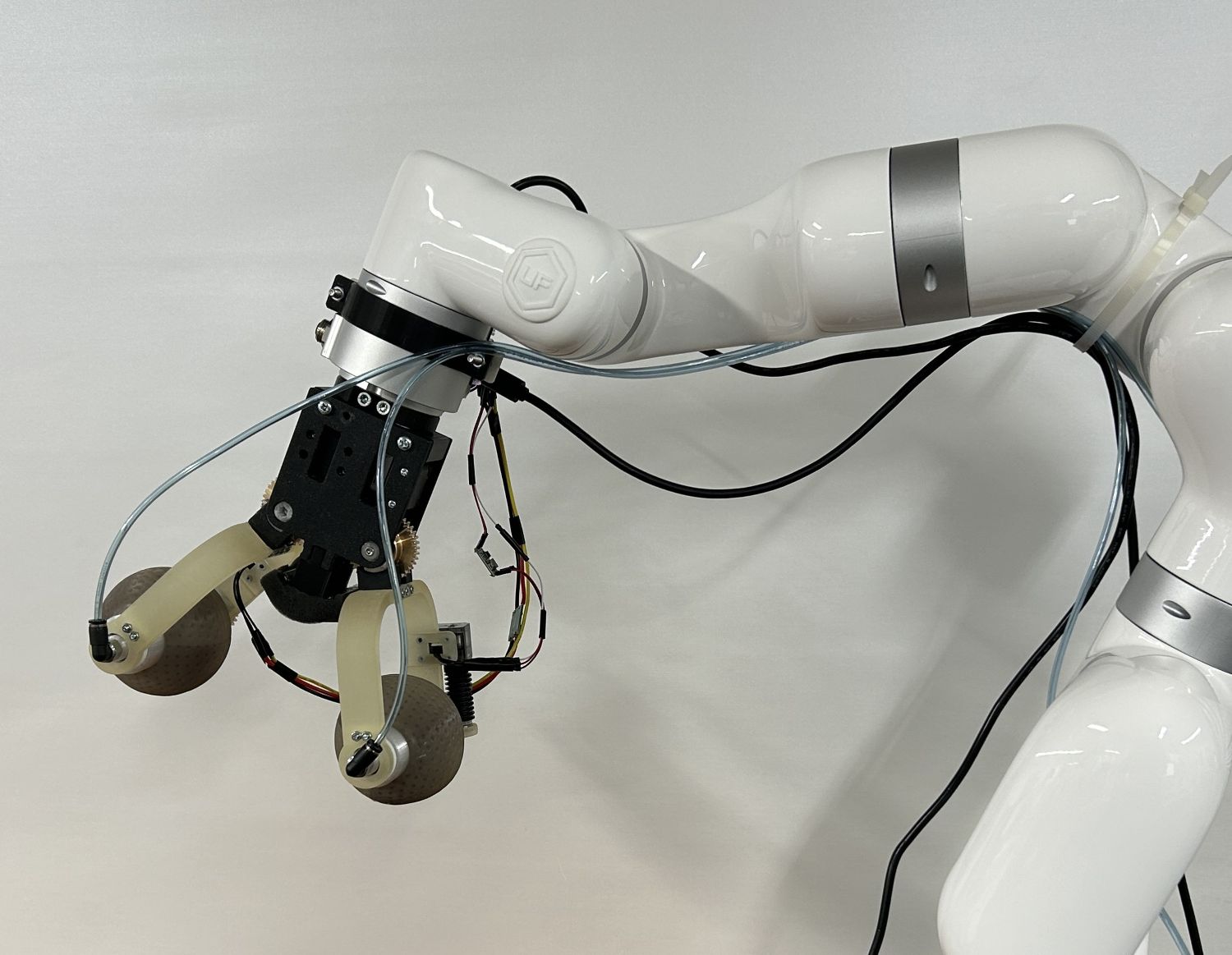}
    \caption{\textbf{Prototype of the proposed roller-jamming gripper mounted on a 7-DoF robot arm (xArm7).}}
    \label{fig:1}
\end{figure}

This paper presents a hybrid roller-jamming gripper in which the same spherical membrane rollers actively draw the object inward after contact and subsequently stiffen under vacuum for post-capture retention, as shown in Fig.~\ref{fig:1}.

The main objectives of this work are as follows:
\begin{enumerate}
    \item A hybrid roller-jamming gripper that integrates active rolling intake and granular-jamming stiffening within the rollers themselves.
    \item A simplified geometric characterisation of the gripper, including a nominal estimate of the minimum supported circular-feature size, together with a bench-level characterisation of the prototype’s force capability.
    \item A controlled experimental evaluation under planar position offsets and combined orientation offsets, together with ablation comparisons that isolate the complementary roles of rolling and jamming.
\end{enumerate}

%%%%%%%%%%%%%%%%%%%%%%%%%%%%%%%%%%%%%%%%%%%%%%%%%%%%%%%%%%%%%%%%%%%%%%%%%%%%%%%%

\section{Gripper Design and System Architecture}

%===============================================================================

\subsection{Overall System Overview}
\begin{figure}[thpb]
  \centering
    \parbox{\columnwidth}{%
      \centering
      \includegraphics[width=\linewidth]{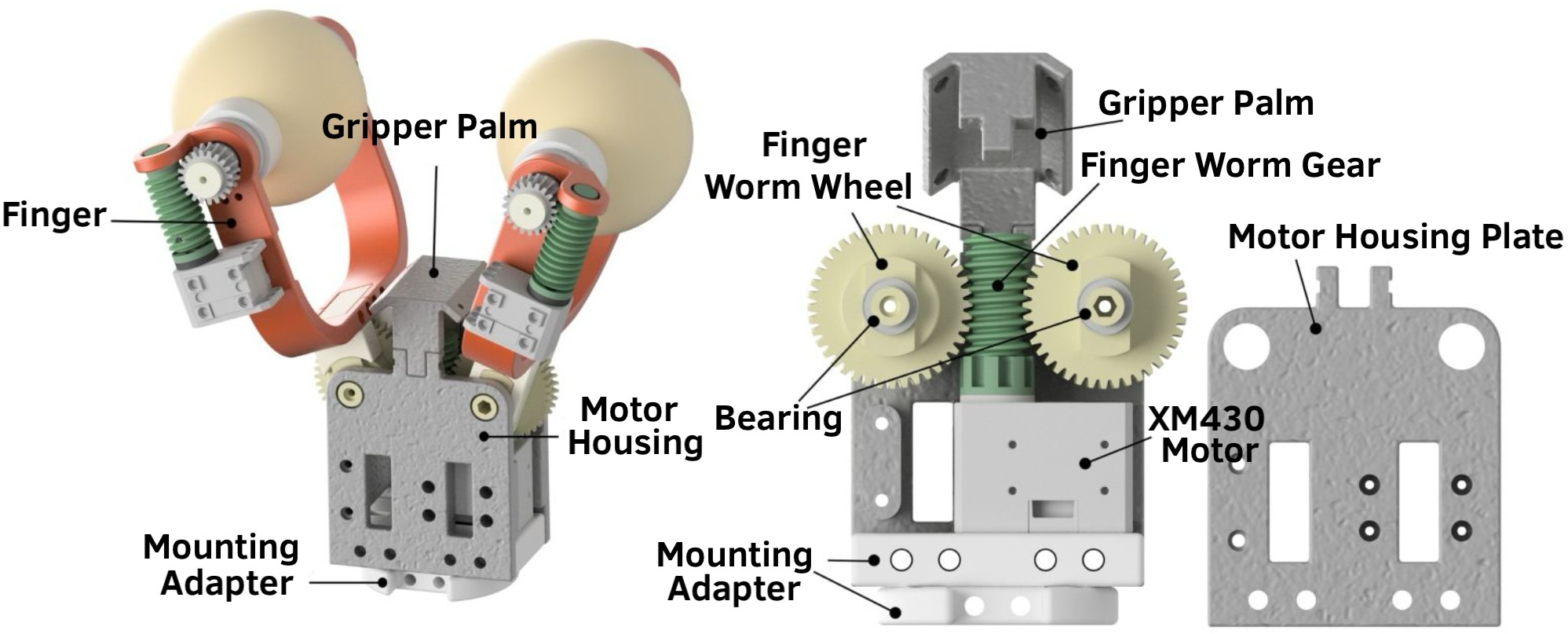}%
  }
    \caption{Mechanical architecture of the proposed gripper: (Left) overall assembly and (Right) base module, showing the gripper palm, mounting adapter, XM430 actuator, and worm-gear finger-closing transmission.}
  \label{fig:2}
\end{figure}

\begin{figure}[thpb]
  \centering
    \parbox{\columnwidth}{%
      \centering
      \includegraphics[width=\linewidth]{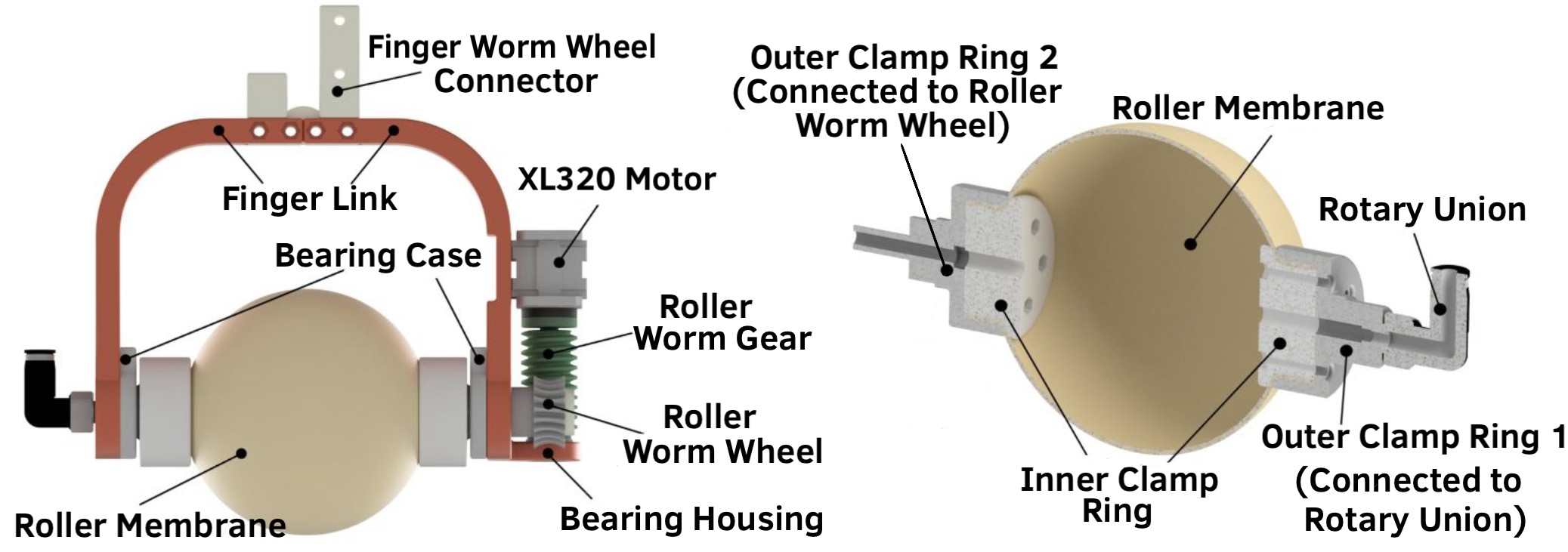}%
  }
  \caption{Roller-finger architecture: (Left) roller-finger module with the XL320-driven worm-gear transmission and (Right) spherical membrane roller construction, including the clamp rings, drive interface, and rotary union for vacuum feedthrough.}
  \label{fig:3}
\end{figure}

The proposed gripper adopts a two-finger architecture composed of a base module and two opposing roller-finger modules, as shown in Fig.~\ref{fig:2} and Fig.~\ref{fig:3}. At the system level, the design combines two functions within the same end-effector: active rolling during acquisition and granular-jamming stiffening during retention. The base module opens and closes both fingers. Each roller-finger module rotates its roller and connects it to the vacuum system for granular jamming. Supporting pneumatic, electronic, and control hardware is kept off-board for experimental evaluation.

%===============================================================================

\subsection{Mechanical Design}
The gripper has three actuated degrees of freedom: one shared DoF for finger opening and closing and one DoF for each roller. The main dimensions and mass of the system are summarised in Table~\ref{tab:mech_specs}.

\begin{table}[thpb]
\centering
\caption{Key mechanical specifications of the gripper.}
\label{tab:mech_specs}
\begin{tabular}{l l}
\hline
\textbf{Item} & \textbf{Value} \\
\hline
Overall envelope (W$\times$L$\times$D) & $70 \times 228 \times 177\,\mathrm{mm}$ \\
Total gripper mass & $1.31\,\mathrm{kg}$ \\
Spherical membrane roller diameter & $\approx 80\,\mathrm{mm}$ \\
Roller axial width & $73\,\mathrm{mm}$ \\
Finger pivot to roller centre & $115.6\,\mathrm{mm}$ \\
Max roller centre-to-centre distance & $285\,\mathrm{mm}$ \\
Finger drive transmission & Worm gear, $1{:}40$ \\
Roller drive transmission (each roller) & Worm gear, $1{:}20$ \\
\hline
\end{tabular}
\end{table}

\subsubsection{Base module}
Finger opening and closing are driven by a single XM430 actuator through a 1:40 worm-gear transmission. The worm engages two worm wheels to move the fingers symmetrically in opposite directions, providing synchronous motion and self-locking. The maximum roller centre-to-centre distance is 285~mm. Because the nominal roller envelopes overlap during closure, the fully closed state is defined by motor current rather than a fixed gap. To protect the transmission and printed couplings, closure is stopped when the XM430 current reaches 0.296~A.

\subsubsection{Roller-finger modules}
Each finger incorporates an independently actuated roller. An XL320 actuator mounted on each finger link drives the roller through a 1:20 worm-gear transmission. This arrangement allows each roller to rotate inward during object intake and outward during release. The roller is supported by bearings at both ends, and each roller has an outer diameter of approximately 80 mm and an axial width of 73 mm.

\subsubsection{Structure and materials}
Most structural parts of the gripper are fabricated by 3D printing in ABS, including the main body components and several transmission supports. An exception is the finger links, which are printed in RGD720 resin because their geometry requires higher dimensional accuracy and finer feature resolution. The motors, worm gears, bearings, fasteners, and rotary unions are off-the-shelf components. Foam pads are attached to the palm to reduce hard contact between the object and the base structure during grasping.
 
\subsubsection{Design choices}
For the roller geometry, cylindrical membrane rollers were initially considered. Preliminary tests showed that the cylindrical membranes folded locally near the central contact region when closing on an object. This prevented smooth deformation around the object surface and reduced contact formation. Spherical membrane rollers were therefore adopted in the final prototype because they produced smoother deformation and more stable contact development.

%===============================================================================

\subsection{Spherical membrane roller construction}

\subsubsection{Membrane fabrication}
\begin{figure}[thpb]
  \centering
    \parbox{\columnwidth}{%
      \centering
      \includegraphics[width=\linewidth]{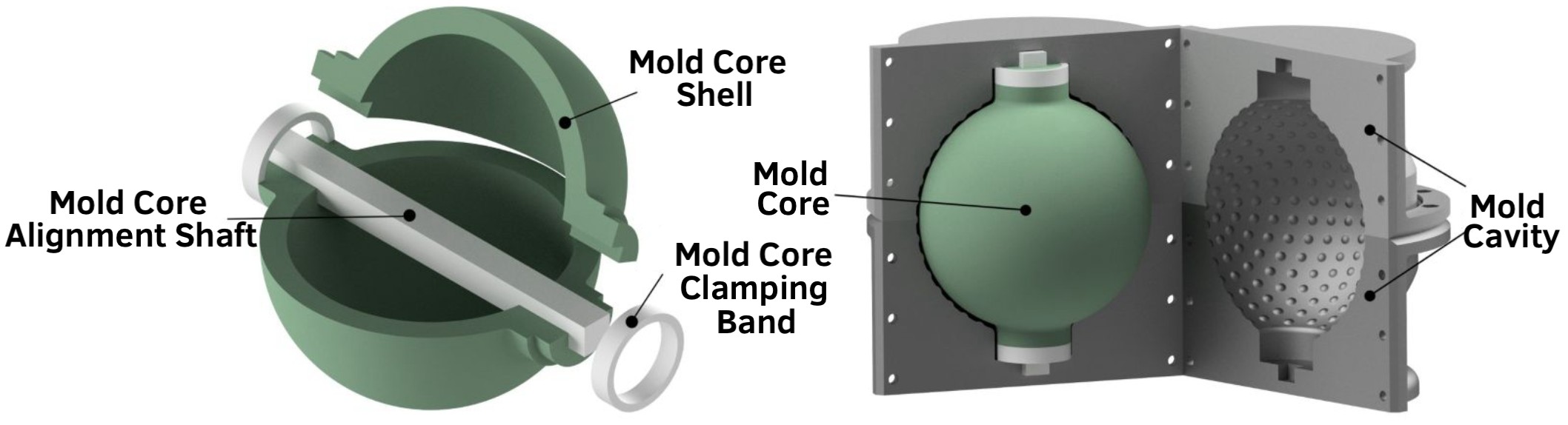}%
  }
  \caption{Membrane mould components: (Left) internal core and alignment shaft and (Right) assembled two-part mould containing the core shell and outer cavity used to form the spherical dimpled membrane.}
  \label{fig:4}
\end{figure}

The membrane is fabricated as a single-piece spherical shell from Dragon Skin 30 silicone, with a nominal thickness of 1 mm and an outer diameter of approximately 80 mm. The membrane is cast in a 3D-printed two-part mould with an internal core, as shown in Fig.~\ref{fig:4}. The gap between the core and cavity determines the membrane thickness. Dimples are distributed over the spherical outer surface. Each dimple has a nominal diameter of 3 mm and height of 0.6 mm. This texture is intended to improve flexibility of the membrane.

\subsubsection{Granular fill}
Each membrane is primarily filled with ground coffee and supplemented with approximately 50 polypropylene (PP) balls of 3~mm and 5.5~mm diameter. This mixed fill was selected to retain the adaptability of the fine granular medium while reducing clumping and preserving airflow during evacuation. The total mass of each roller-finger module is approximately 0.12 kg.

\subsubsection{End sealing and vacuum feedthrough}
Both ends of the membrane are terminated by 3D-printed ABS clamp rings that compress the membrane to form an airtight seal. One end provides the mechanical interface to the roller drive, while the opposite end connects to a pneumatic rotary union, allowing vacuum to be applied while the roller remains free to rotate.

\subsubsection{Serviceability and durability}
No membrane failures were observed during approximately 1000 grasp trials conducted using the same membrane set. Because the membrane is secured by clamped end rings rather than permanent bonding, the membrane and infill can be replaced in approximately 10 minutes.

%===============================================================================

\subsection{SYSTEM INTEGRATION: PNEUMATICS, ELECTRONICS, AND ROBOT-ARM PLATFORM}
\begin{figure}[thpb]
  \centering\centering
      \includegraphics[width=\linewidth]{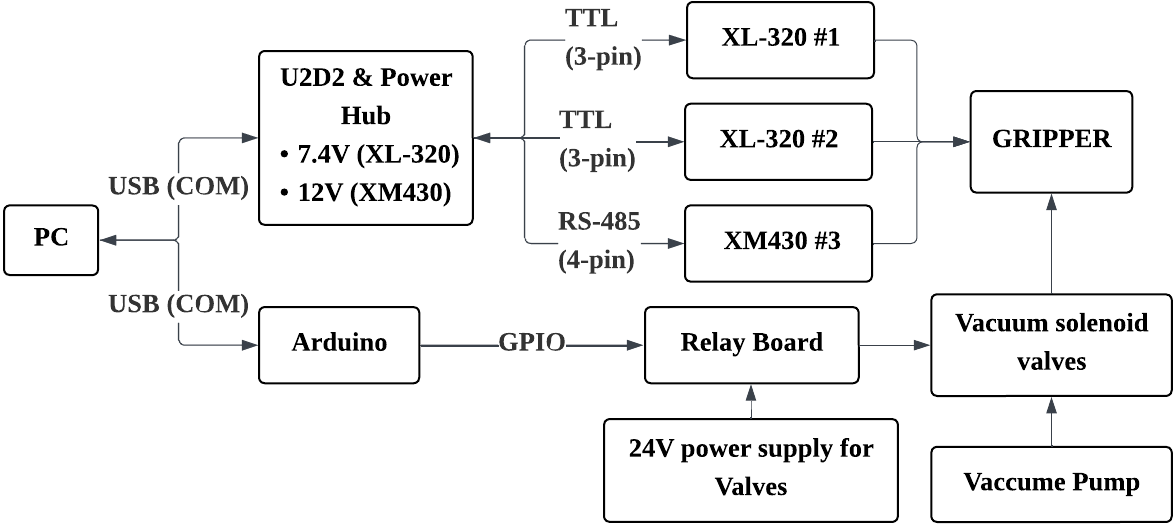}
  \caption{Control, communication, electrical-power, and pneumatic architecture of the prototype gripper system.}
  \label{fig:5}
\end{figure}

During the experiments, the gripper was mounted on a 7-DoF xArm7 through a custom flange. The vacuum pump, valves, controllers, and power supplies remained off-board. The gripper is not specific to the xArm7, which was used only to position and move the gripper during the experiments. A host PC coordinated the robot arm, the three Dynamixel actuators, and the Arduino-based valve controller, as shown in Fig.~\ref{fig:5}.

The three gripper actuators communicated through a U2D2 interface. The XM430 finger-closing actuator was supplied at 12~V, while the two XL320 roller actuators were supplied at 7.4~V through a DC--DC converter. Vacuum was provided by an external rotary-vane pump with a free-air displacement of 42~L/min and routed through two independently controlled solenoid valves. The valves were relay-driven by the Arduino and supplied at 24~V. A pressure gauge installed between the valves and the roller rotary unions measured a vacuum level of $-0.082$~MPa during jamming.

%%%%%%%%%%%%%%%%%%%%%%%%%%%%%%%%%%%%%%%%%%%%%%%%%%%%%%%%%%%%%%%%%%%%%%%%%%%%%%%%

\section{GEOMETRIC ANALYSIS AND BENCH CHARACTERISATION}

%===============================================================================

\subsection{Nominal Closing Geometry}
\begin{figure}[thpb]
  \centering\centering
      \includegraphics[width=0.8\linewidth]{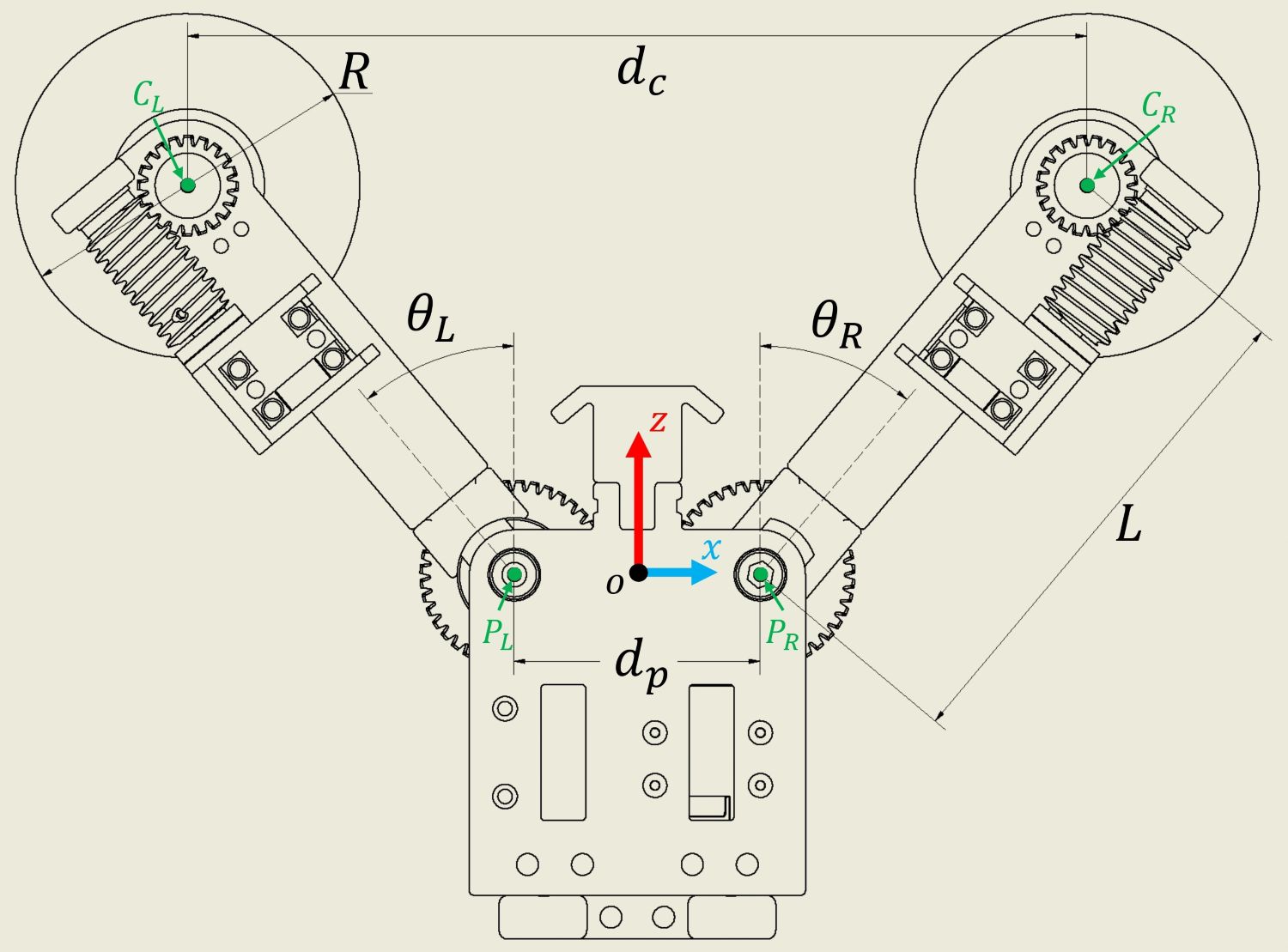}
  \caption{Simplified front-view rigid-envelope model of the gripper. The fingers rotate symmetrically about pivots separated by $d_p$, with pivot-to-roller-centre distance $L$, roller radius $R$, and roller-centre spacing $d_c$.}
  \label{fig:6}
\end{figure}

In the simplified front-view model shown in Fig.~\ref{fig:6}, the gripper is represented by two rigid links of length $L$. The links rotate symmetrically about pivots separated by $d_p$. Each spherical membrane roller is approximated by an undeformed circular envelope of radius $R$. For a symmetric finger angle $\theta$, the roller-centre spacing is

\begin{equation}
d_c(\theta)=d_p+2L\sin\theta ,
\label{eq:roller_spacing}
\end{equation}

and the nominal inter-roller gap is

\begin{equation}
g(\theta)=d_c(\theta)-2R .
\label{eq:roller_gap}
\end{equation}

Here, $g(\theta)>0$ represents a finite gap, $g(\theta)=0$ represents tangential contact between the nominal roller envelopes, and $g(\theta)<0$ represents geometric overlap. In the physical
prototype, the membrane rollers deform during contact, and closure is terminated using a motor-current threshold rather than a fixed geometric gap. This rigid-envelope model describes only the nominal closing geometry. Note that this model excludes membrane deformation, granular redistribution, friction, asymmetric contact, and three-dimensional effects. It should therefore not be interpreted as an accurate prediction of grasp success.

%===============================================================================

\subsection{Nominal Minimum Supported Circular-Feature Size}
\begin{figure}[thpb]
  \centering
    \parbox{\columnwidth}{%
      \centering
      \includegraphics[width=0.65\linewidth]{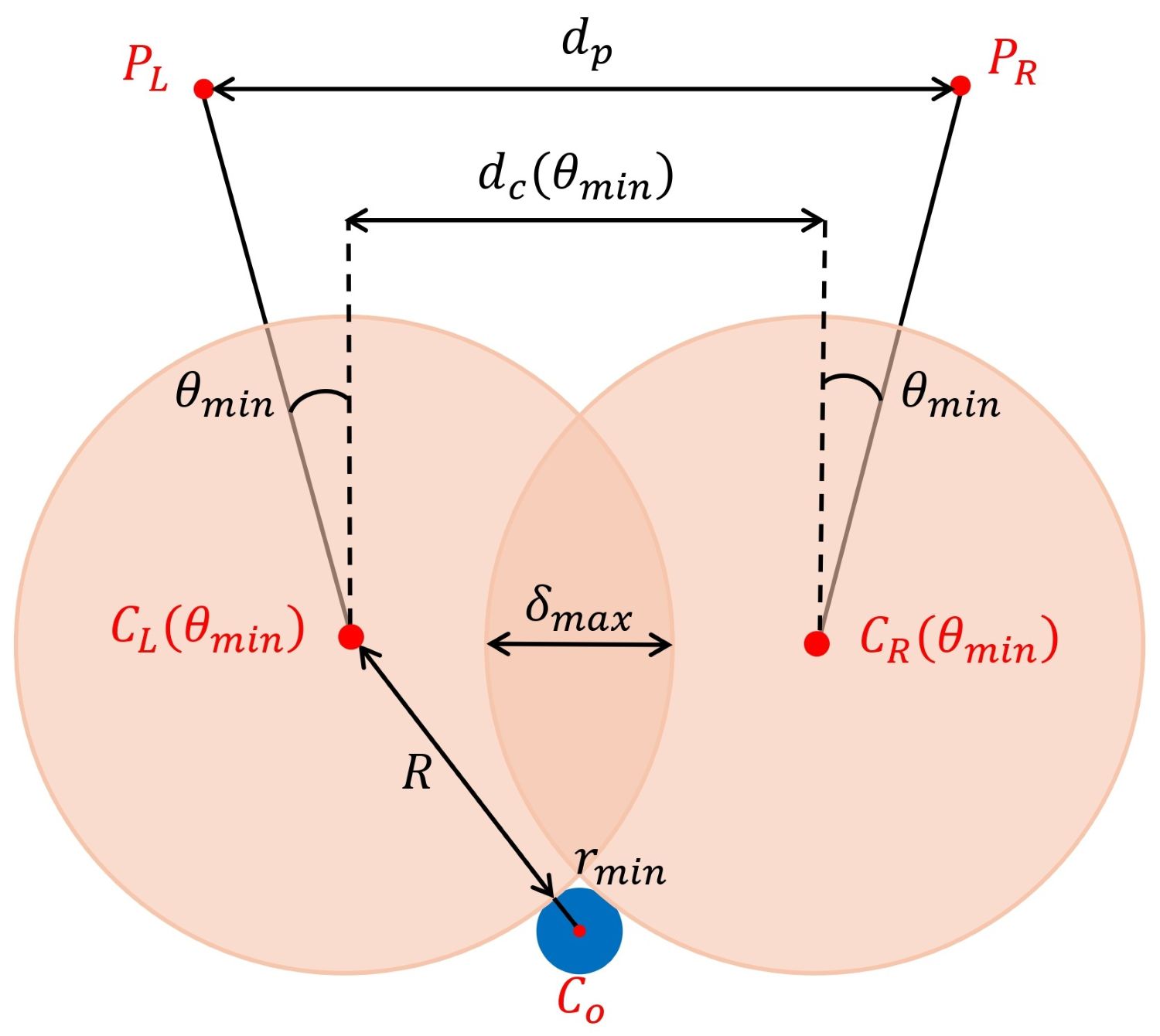}%
    }%
  \caption{Nominal fully closed geometry used to estimate the minimum supported circular-feature radius $r_{\min}$. The undeformed roller envelopes have radius $R$ and centre-to-centre spacing $d_c(\theta_{\min})$.}
  \label{fig:7}
\end{figure}

To obtain a nominal geometric reference for small-object enclosure, the gripper is considered at its minimum finger angle $\theta_{\min}$, as shown in Fig.~\ref{fig:7}. The two undeformed roller envelopes have radius $R$ and centre-to-centre spacing $d_c(\theta_{\min})$. The supported feature is approximated as a circle that is tangent to both roller envelopes and to the support plane. Applying the corresponding tangency conditions gives the minimum supported circular-feature radius as

\begin{equation}
r_{\min}
=
\frac{d_c^2(\theta_{\min})}{16R}.
\label{eq:minimum_feature}
\end{equation}

The nominal overlap between the two roller envelopes at this configuration is $\delta_{\max}=2R-d_c(\theta_{\min})$. For the current prototype, $R=40$ mm and $d_c(\theta_{\min})=35$ mm. Equation~\eqref{eq:minimum_feature} therefore gives $r_{\min}\approx1.9~\mathrm{mm}$, corresponding to a nominal diameter of approximately $3.8$ mm. This value is a geometric reference rather than an experimentally verified graspability limit. Membrane deformation, granular redistribution, contact asymmetry, friction, and surface geometry may increase or decrease the minimum feature size achievable in practice.

%===============================================================================

\subsection{Effect of Front-View Tilt on Minimum Supported Circular-Feature Size}
To examine the nominal effect of front-view tilt, the fully closed roller pair is rotated by an angle $\Delta\theta$ while maintaining the roller-centre spacing $d=d_c(\theta_{\min})$. The left roller and the circular feature are assumed to remain tangent to the support plane. The feature is also assumed to remain tangent to both undeformed roller envelopes. Under these assumptions, the minimum supported circular-feature radius is

\begin{equation}
\begin{aligned}
r_{\min}(\Delta\theta)=
\left(
\frac{
\sqrt{4R + 2d\sin\Delta\theta}
- 2\sqrt{R}\cos\Delta\theta
}{
2\sin\Delta\theta
}
\right)^2.
\end{aligned}
\label{eq:tilted_min_feature}
\end{equation}

As $\Delta\theta \rightarrow 0$, this expression approaches $d^2/(16R)$, recovering the zero-tilt result in \eqref{eq:minimum_feature}. Equation~\eqref{eq:tilted_min_feature} describes only an idealised two-dimensional geometric trend. It does not account for membrane deformation, asymmetric or out-of-plane contact, or the full roll--pitch--yaw orientation offsets used in the experiments.

\begin{figure}[thpb]
  \centering
    \parbox{\columnwidth}{%
      \centering
      \includegraphics[width=0.8\linewidth]{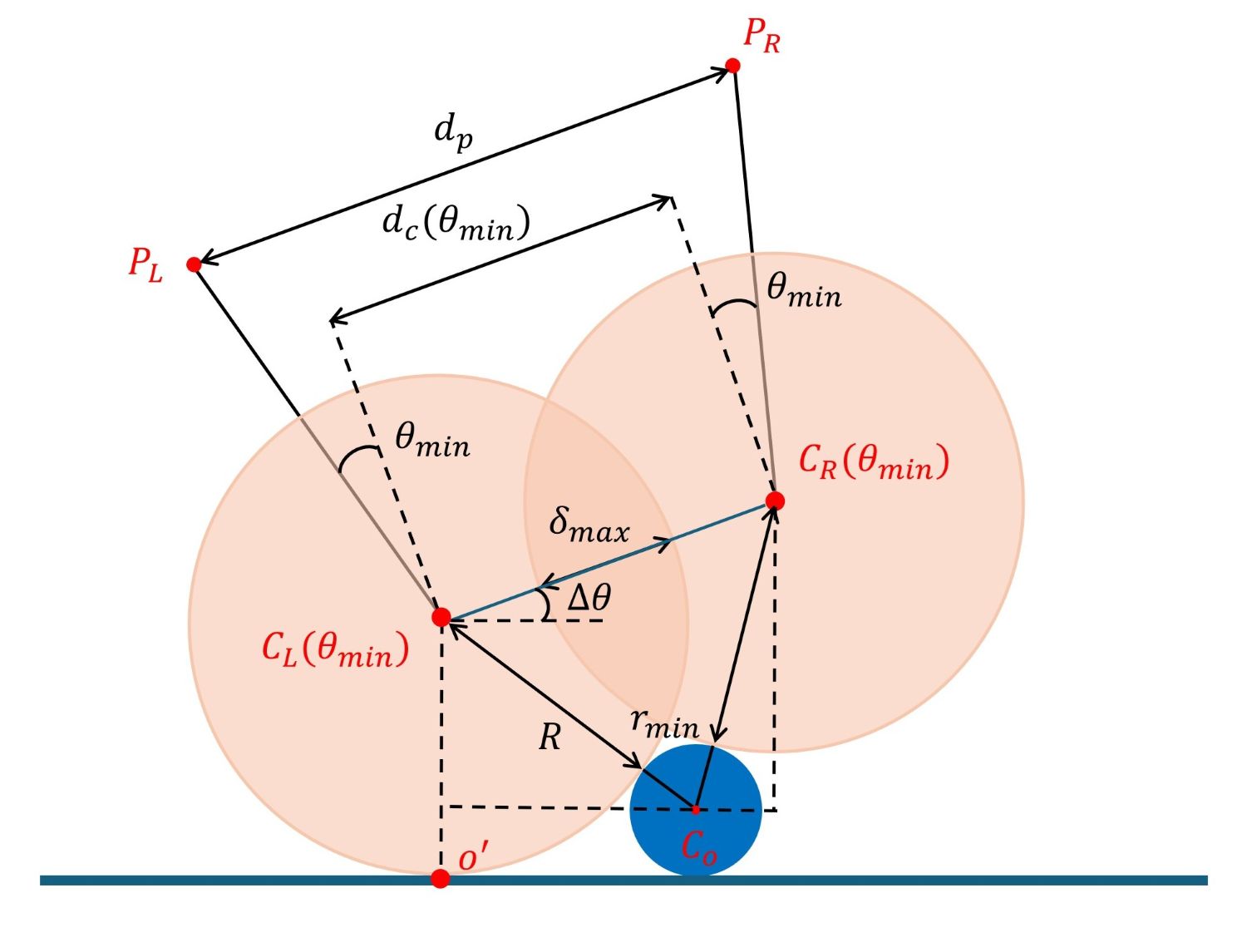}%
    }%
    \caption{Simplified front-view tilted-contact geometry used to estimate the minimum supported circular-feature radius. The roller pair is tilted by {\unboldmath$\Delta\theta$} while maintaining the nominal centre spacing $d_c(\theta_{\min})$. The supported feature remains tangent to both
    undeformed roller envelopes and the support plane.}
  \label{fig:8}
\end{figure}

%===============================================================================

\subsection{Bench Characterisation}
Three types of preliminary bench test were conducted to characterise the prototype: roller traction, grasp slip-out resistance, and clamping force. Each test was performed once; therefore, the reported values represent single-trial maximum measurements rather than statistical force estimates. An S-type load cell (LC-1122, 500~N) connected to an HX711 amplifier was used for all measurements. This type of bench-level evaluation is consistent with prior work that emphasises  explicit external performance metrics for robot end-effectors \cite{falco2020benchmarking}.

For the roller traction test, the load cell was connected by a string to a 3D-printed spherical target. The rollers were driven inward with vacuum disabled, and the maximum tensile force
measured before slip was 8.4~N. For the grasp slip-out test, the same spherical target was secured using the complete grasping sequence described in Section~\ref{subsec:IVB}. The xArm7 then moved vertically upward until the target slipped from the grasp. The maximum measured slip-out force was 25.6~N. For the clamping-force test, the load cell was positioned directly between the fingers and the gripper was progressively closed. The maximum recorded compressive force was 27.7~N, at which point the 3D-printed mechanical parts failed. This value represents the structural limit observed during the test rather than a rated operating clamping force.

\begin{figure}[thpb]
  \centering
    \parbox{\columnwidth}{%
      \centering
      \includegraphics[width=\linewidth]{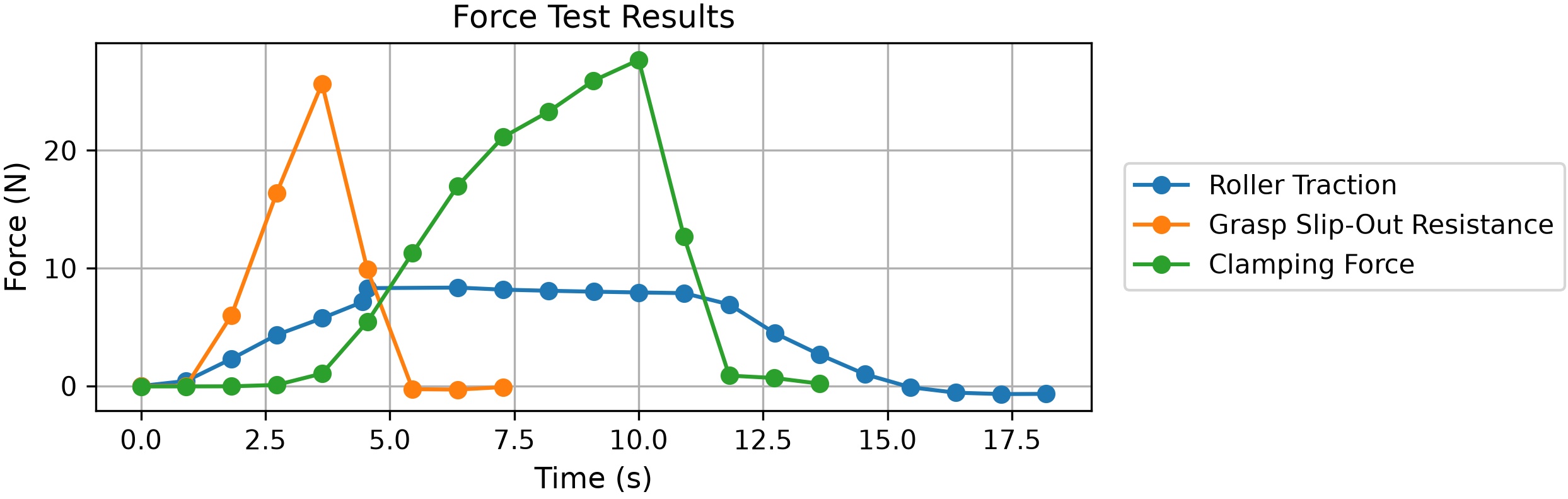}%
    }%
  \caption{Preliminary bench force tests. Force time histories from the single roller-traction, grasp slip-out-resistance, and clamping-force trials. The maximum measured forces were 8.4, 25.6, and 27.7~N, respectively.}
  \label{fig:9}
\end{figure}

%%%%%%%%%%%%%%%%%%%%%%%%%%%%%%%%%%%%%%%%%%%%%%%%%%%%%%%%%%%%%%%%%%%%%%%%%%%%%%%%

\section{Control and Grasp Sequence}

\subsection{Finite State Machine Control}
\begin{figure}[thpb]
  \centering\centering
      \includegraphics[width=\linewidth]{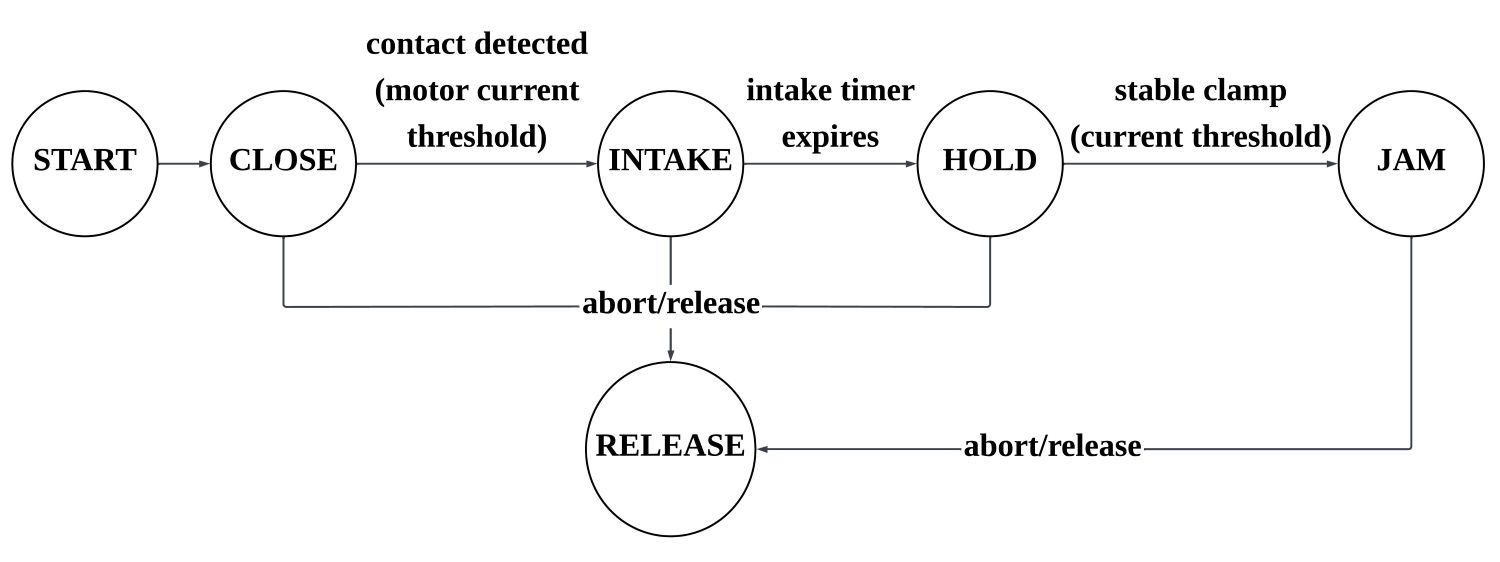}
  \caption{Finite-state machine of the grasp controller. \texttt{CLOSE} and \texttt{HOLD} use motor-current and dwell conditions, whereas \texttt{INTAKE} and \texttt{JAM} use prescribed durations. Abort or manual-release commands return the system to \texttt{RELEASE}.}
  \label{fig:10}
\end{figure}

The gripper is controlled by a Python-based finite-state machine (FSM) running on the host PC at 50 Hz. The FSM comprises five states: \texttt{CLOSE}, \texttt{INTAKE}, \texttt{HOLD}, \texttt{JAM}, and \texttt{RELEASE}, as shown in Fig.~\ref{fig:10}.

%===============================================================================

\subsection{State Definitions}
\label{subsec:IVB}
The role and transition condition of each state are summarised below:
\begin{itemize}
    \item \texttt{CLOSE:}
    The controller switches to \texttt{INTAKE} when the motor current exceeds the object-specific \texttt{CLOSE} threshold for the prescribed dwell time. This current increase is used as an indirect indication of initial contact.
    
    \item \texttt{INTAKE:}
    The two XL320-driven rollers rotate inward for the object-specific duration listed in Table~\ref{tab:object_parameters}. This draws the object toward the gripper centre, which increases membrane engagement before jamming.
    
    \item \texttt{HOLD:}
    The fingers increase the clamping load until the motor current exceeds the object-specific \texttt{HOLD} threshold. The controller will switch to \texttt{JAM} once the threshold is maintained for the prescribed dwell time.
    
    \item \texttt{JAM:}
    Vacuum is applied to both spherical membrane rollers for a fixed 2-s interval to induce granular jamming. Lifting and transport begin after this interval.
    
    \item \texttt{RELEASE:}
    The vacuum lines are vented, the rollers rotate outward, and the fingers open to return the gripper to its initial configuration. This state is also entered after an abort or manual-release command.
\end{itemize}

The motor current had to remain above the \texttt{CLOSE} and \texttt{HOLD} thresholds for 0.5 and 0.8~s, respectively.

%%%%%%%%%%%%%%%%%%%%%%%%%%%%%%%%%%%%%%%%%%%%%%%%%%%%%%%%%%%%%%%%%%%%%%%%%%%%%%%%

\section{Experimental Protocol}

%===============================================================================

\subsection{Goal and Evaluation Scope}
The experiments are designed to examine the mechanism-level behaviour of the proposed gripper under imperfect first contact. The two test protocols are the planar-offset test and the orientation-offset test. In each case, the objective is to determine whether the gripper can still acquire the object and maintain the grasp through a fixed retention routine. The evaluation therefore distinguishes between acquisition and post-capture retention, rather than attempting to validate a complete household-manipulation system. Ablation trials were also conducted on a subset of objects to examine the separate contributions of rolling intake and jamming.

\begin{figure}[thpb]
  \centering\centering
      \includegraphics[width=0.8\linewidth]{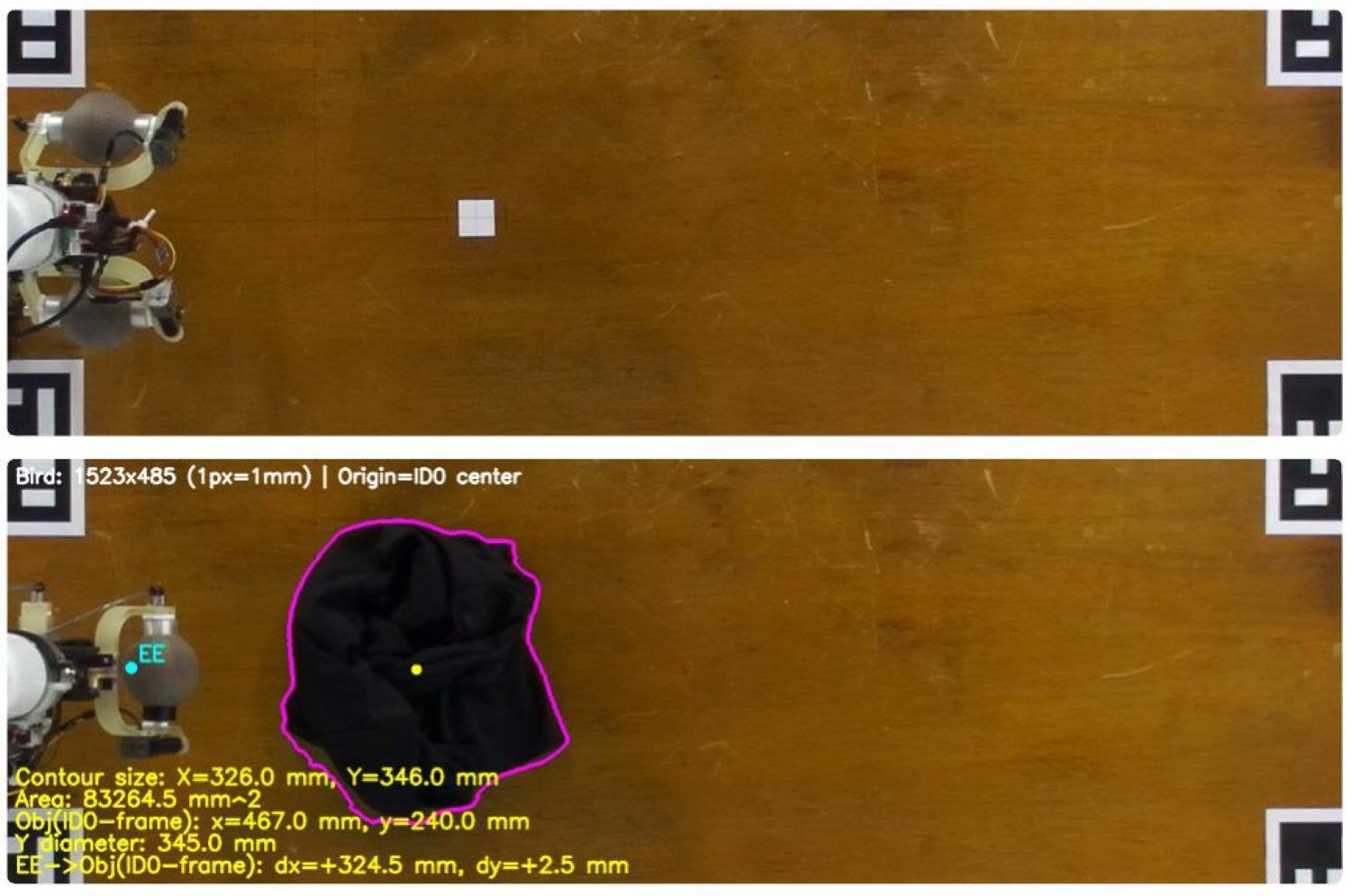}
    \caption{Experimental setup (top) and example of camera-based object localisation (bottom). For the T-shirt, pressure gauge, and cable, the overhead camera estimates the object centroid on the tabletop and in-plane extents after each manual reset. AprilTags define the tabletop reference frame used for localisation and offset generation.}
  \label{fig:11}
\end{figure}

%===============================================================================

\subsection{Experimental Setup}
All grasp attempts use the same top-down grasping strategy. Objects are placed manually and reset between trials. For the pen, marble, gallon jug, plastic cup, and paper sheet, the in-plane dimensions and nominal geometric centre were determined manually before the experiments. These measurements were kept fixed and were not recalculated after each manual reset. After each trial, these objects were manually repositioned using markings on the tabletop corresponding to their nominal placement. For the T-shirt, pressure gauge, and cable, the overhead ZED Mini camera estimated the centroid and in-plane dimensions after each reset. Manual resetting could not reliably reproduce their position or in-plane outline. AprilTags defined the tabletop reference frame used for camera-based localisation and offset generation, as shown in Fig.~\ref{fig:11}.

%===============================================================================

\subsection{Test Objects}
\begin{figure}[thpb]
  \centering
      \centering
      \includegraphics[width=0.9\linewidth]{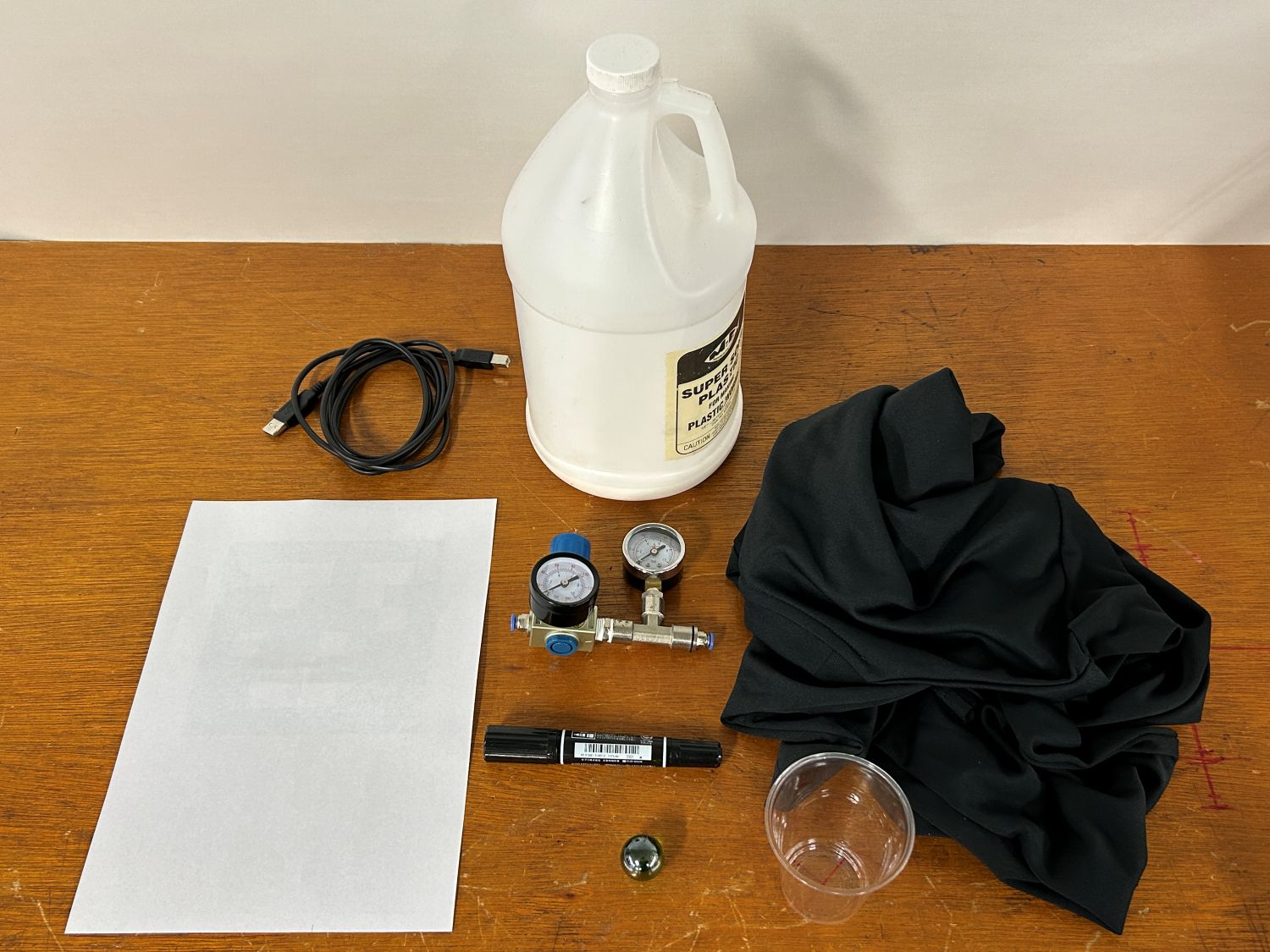}%
  \caption{Test objects used for evaluation: pen, marble, gallon jug, plastic cup, paper sheet, T-shirt, pressure gauge, and cable}
  \label{fig:12}
\end{figure}

Eight objects were evaluated: a pen, marble, gallon jug, plastic cup, paper sheet, T-shirt, pressure gauge, and cable, as shown in Fig.~\ref{fig:12}. They were selected to represent different grasping challenges, including slender, spherical, flat, deformable, thin-walled, textile, and irregular objects. Their main properties and roles in the evaluation are summarised in Table~\ref{tab:test_objects}. This property-based selection follows previous benchmarking studies~\cite{matheus2010benchmarking}, but the objects are not taken from a standardised set such as YCB~\cite{calli2015ycb}. The experiments therefore cover representative contact conditions. They do not isolate individual object properties or provide direct comparisons with other grippers.

\begin{table*}[thpb]
\caption{Physical characteristics and contact challenges represented by the test objects.}
\label{tab:test_objects}
\centering
\footnotesize
\setlength{\tabcolsep}{5pt}
\renewcommand{\arraystretch}{1.12}
\begin{tabular}{
    p{0.12\linewidth}
    c
    p{0.31\linewidth}
    p{0.42\linewidth}
}
\hline
Object & Mass [g] & Physical characteristics & Contact challenge represented \\
\hline

Pen
& 24.6
& Rigid, slender, and low-profile
& Shallow contact with a narrow object. \\

Marble
& 19.9
& Rigid, smooth, small, and spherical
& Localised contact and enclosure of a small sphere. \\

Gallon jug
& 420
& Rigid, large, and relatively heavy
& Retention of a large object under higher loading. \\

Plastic cup
& 6.0
& Lightweight, thin-walled, and deformable
& Grasping without excessive deformation. \\

Paper sheet
& 4.2
& Thin, flat, flexible, and low-profile
& Shallow and potentially one-sided contact. \\

T-shirt
& 180
& Highly deformable textile
& Changing shape and contact region during acquisition. \\

Pressure gauge
& 290
& Rigid and geometrically irregular
& Orientation-dependent contact on a complex shape. \\

Cable
& 70
& Flexible, elongated, and low-profile
& Gathering an object from partial initial contact. \\

\hline
\end{tabular}
\end{table*}

%===============================================================================

\subsection{Calibrated Experimental Parameters}
Several grasp parameters were calibrated separately for each object before the repeated experiments, as summarised in Table~\ref{tab:object_parameters}. Here, grasp height is the commanded vertical position of the gripper tool centre point (TCP) along the tabletop-frame $z$-axis. The reference ($z=0$) is defined as the position at which the tip of the membrane just touches the tabletop, while positive $z$ values move the gripper upward. For each protocol, the grasp height was set to the lowest collision-free position. At this height, both rollers provided sufficient contact and the membranes could deform around the intended grasp region. The intake duration and HOLD setting were selected to provide sufficient roller engagement and clamping without excessive object deformation.

The planar-offset and ablation tests used the same calibrated grasp height. A separate height was used for the orientation-offset tests because tilted approaches required additional table clearance. Once selected, all object-specific parameters were fixed and were not changed between offset conditions or repeated trials.

\begin{table*}[thpb]
\centering
\caption{Object-specific parameters used in the repeated experiments.}
\label{tab:object_parameters}
\begin{tabular}{lccccc}
\hline
Object &
\begin{tabular}[c]{@{}c@{}}Planar/ablation\\grasp height [mm]\end{tabular} &
\begin{tabular}[c]{@{}c@{}}Orientation\\grasp height [mm]\end{tabular} &
\begin{tabular}[c]{@{}c@{}}Intake\\duration [s]\end{tabular} &
\begin{tabular}[c]{@{}c@{}}CLOSE current\\threshold [mA]\end{tabular} &
\begin{tabular}[c]{@{}c@{}}HOLD current\\threshold [mA]\end{tabular} \\
\hline
Pen            & 6 & 6 & 5 & 40 & 200 \\
Marble         & 10 & 10 & 5 & 130 & 230 \\
Gallon jug     & 254 & 254 & 2 & 80 & 230 \\
Plastic cup    & 58 & 58 & 4 & 40 & 200 \\
Paper sheet    & -10 & -10 & 5 & 70 & 70 \\
T-shirt        & 0 & 6 & 8 & 40 & 200 \\
Pressure gauge & -5 & 6 & 5 & 130 & 230 \\
Cable          & 0 & 6 & 8 & 50 & 200 \\
\hline
\end{tabular}
\end{table*}

%===============================================================================

\subsection{Success Metric and Retention Routine}
A trial was classified as successful if the object was acquired, lifted clear of the table, and retained throughout the complete post-grasp motion routine. All other outcomes were classified as non-successes for the success-rate analysis. The success rate for each test protocol was calculated as the number of successful trials divided by the total number of repeated trials.

Retention was evaluated using a fixed motion sequence relative to the post-lift pose. The sequence comprised translations of $\pm75$~mm along $x$ and $y$ axes, translations of $\pm30$~mm along $z$, and rotations of $\pm15^\circ$ in yaw, pitch, and roll. The same retention routine and motion order are used across all experiments. Note that the $z$-axis translations in this routine are applied after the object has already been acquired and lifted. They are therefore post-grasp disturbance motions and are distinct from the commanded grasp height used during acquisition.

%===============================================================================

\subsection{Planar Offset Robustness Test}
To evaluate robustness to planar offsets, two-dimensional offsets are applied on a $3 \times 3$ grid centred on the estimated geometric centre of the object in the tabletop frame. Let \(L_x\) and \(L_y\) denote the object extents along the tabletop \(x\) and \(y\) axes, respectively. For the pen, marble, gallon jug, plastic cup, and paper sheet, these extents were obtained from fixed manual measurements. For the T-shirt, pressure gauge, and cable, they were estimated by the overhead camera after each manual reset. The maximum offset magnitude along each axis is set to 20\% of the corresponding extent,
\begin{equation}
\Delta_x^{\max} = 0.2L_x,\qquad \Delta_y^{\max} = 0.2L_y .
\end{equation}
The applied offsets are then sampled as
\begin{equation}
\delta_x \in \{-\Delta_x^{\max},\,0,\,\Delta_x^{\max}\}, \qquad
\delta_y \in \{-\Delta_y^{\max},\,0,\,\Delta_y^{\max}\},
\end{equation}
yielding nine planar-offset conditions per object. Each condition was repeated three times for each of the eight objects, resulting in $8 \times 9 \times 3 = 216$ planar-offset trials. The calibrated planar grasp height listed in Table~\ref{tab:object_parameters} was used for all nine offset conditions and all three repeated trials of the corresponding object.

%===============================================================================

\subsection{Orientation Offset Robustness Test}
To evaluate robustness to approach orientation offsets, roll, pitch, and yaw offsets were applied at $\{-10^{\circ},0^{\circ},+10^{\circ}\}$ across the 26 non-zero combinations. The all-zero condition was excluded because it represents the nominal orientation. Each orientation condition was repeated three times for every object, resulting in 78 orientation trials per object and $8 \times 26 \times 3 = 624$ orientation-offset trials in total.

For the paper sheet, the offset magnitude was reduced to $\{-5^{\circ},0^{\circ},+5^{\circ}\}$ to avoid gripper-table collision during the low-height approach required for acquisition. The same 26 non-zero orientation combinations and three repetitions per condition were used.

Because roll and pitch offsets reduce the available table clearance during tilted approaches, the orientation-test grasp height was calibrated separately where required. The values reported in Table~\ref{tab:object_parameters} were kept fixed across all 26 orientation conditions and all three repetitions of each object.

%===============================================================================

\subsection{Comparison Conditions}
The ablation tests used the pen, gallon jug, and cable. In the roller-only condition (JAM OFF), jamming was disabled, whereas in the jamming-only condition (ROLLER OFF), roller intake was disabled. All other trial settings were unchanged, including object placement, grasp execution, parameter settings, and the retention routine. The ablation trials used the same calibrated non-tilted grasp height as the planar-offset protocol for the corresponding object. Each of the nine planar-offset conditions was repeated three times under both ablation modes for each of the three selected objects. The ablation evaluation therefore comprised $3 \times 9 \times 3 \times 2 = 162$ trials.

%%%%%%%%%%%%%%%%%%%%%%%%%%%%%%%%%%%%%%%%%%%%%%%%%%%%%%%%%%%%%%%%%%%%%%%%%%%%%%%%

\section{Results}

\subsection{Overall Performance Summary}
\begin{figure}[thpb]
  \centering
      \centering
      \includegraphics[width=\linewidth]{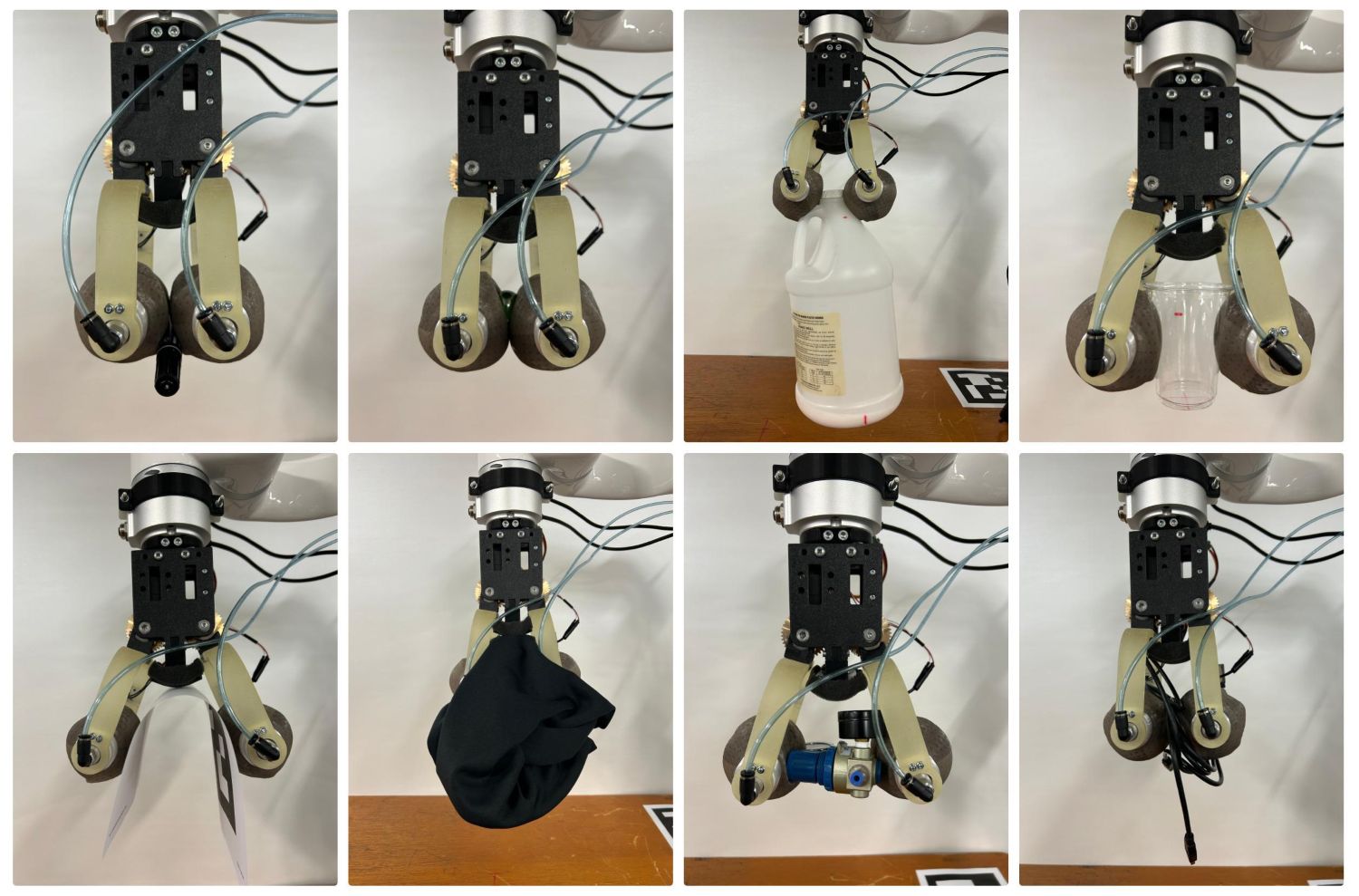}%
  \caption{Successful grasps with the roller-jamming gripper on the eight test objects used in evaluation. Top row (left to right): pen, marble, gallon jug, plastic cup. Bottom row (left to right): paper sheet, T-shirt, pressure gauge, cable.}
  \label{fig:13}
\end{figure}

Table~\ref{tab:repeated_success_summary} summarises the repeated-trial results. The hybrid gripper achieved 812 successes in 840 trials across the main evaluation. Disabling either rolling or jamming substantially reduced performance in the three-object ablation evaluation, indicating that the two mechanisms provide complementary functions. Representative successful grasps on the eight test objects are shown in Fig.~\ref{fig:13}.

\begin{table}[thpb]
    \centering
    \caption{Success rate across the test protocols with two-sided 95\% Wilson score confidence intervals.}
    \label{tab:repeated_success_summary}
        \begin{tabular}{lccc}
        \hline
        Test protocol & Successful trials & Success rate & 95\% CI \\
        \hline
        Planar offset
            & 215/216 & 99.5\% & 97.4--99.9\% \\
        Orientation offset
            & 597/624 & 95.7\% & 93.8--97.0\% \\
        Main evaluation
            & 812/840 & 96.7\% & 95.2--97.7\% \\
        \hline
        JAM OFF
            & 54/81 & 66.7\% & 55.9--76.0\% \\
        ROLLER OFF
            & 24/81 & 29.6\% & 20.8--40.3\% \\
        \hline
        \end{tabular}
\end{table}

%===============================================================================

\subsection{Planar Offset Robustness Results}
\begin{figure*}[thpb]
  \centering\centering
      \includegraphics[width=\linewidth]{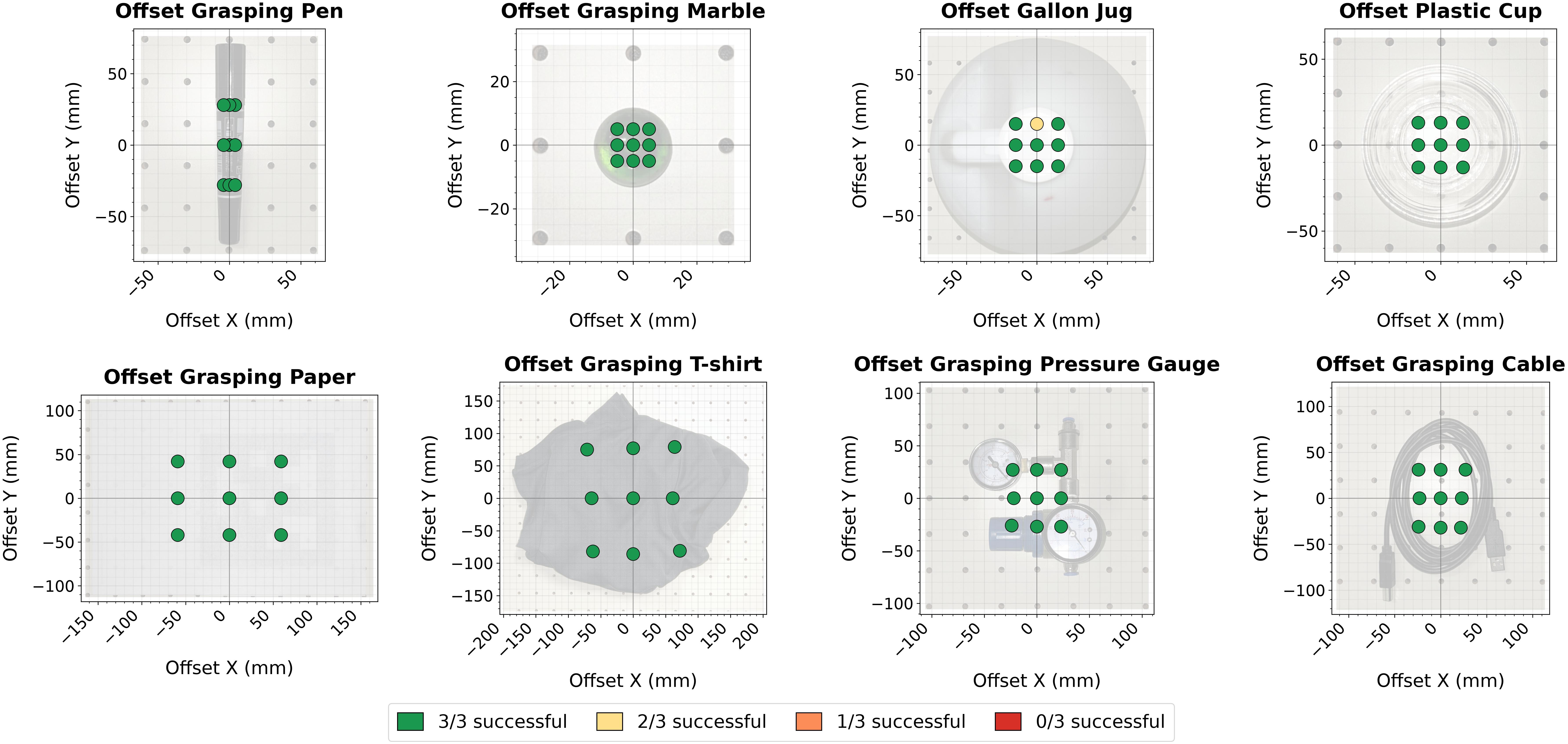}
  \caption{Planar-offset test results for the eight objects, with three trials per condition. Each coloured marker corresponds to one offset condition repeated three times; the marker colour indicates the number of successful trials out of three. Offsets are applied on a $3 \times 3$ grid about the estimated object centre and scaled to $\pm 20\%$ of the corresponding in-plane object extents.}
  \label{fig:14}
\end{figure*}

Fig.~\ref{fig:14} summarises the repeated planar-offset results. Seven of the eight objects completed all 27 trials successfully across the nine tested offset conditions. The gallon jug completed 26 of 27 trials successfully and accounted for the only failure in the planar-offset evaluation. The overall planar-offset success rate was 215/216, or 99.5\%. These results show consistent performance across the three repeated trials under the tested planar offsets. However, the result should be interpreted within the controlled tabletop protocol and the object-specific parameter calibration.

%===============================================================================

\subsection{Planar-Offset Ablation Results}
\begin{figure}[thpb]
  \centering
      \centering
      \includegraphics[width=\linewidth]{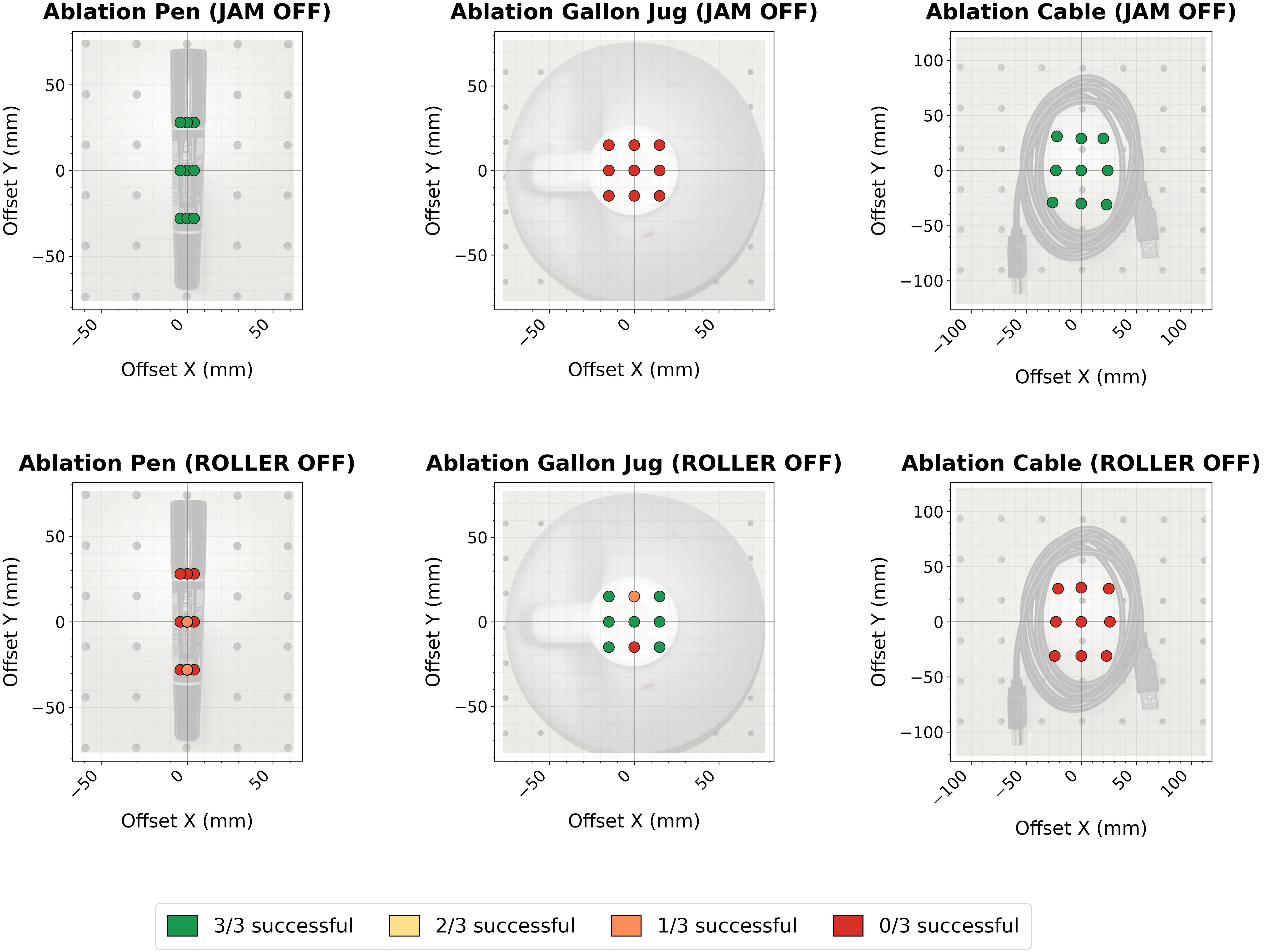}%
    \caption{Ablation results for the pen, gallon jug, and cable under the planar-offset protocol. The top row shows the roller-only condition (JAM OFF), and the bottom row shows the jamming-only condition (ROLLER OFF). Marker colour indicates the number of successful trials out of three at each offset.}
  \label{fig:15}
\end{figure}

Across the corresponding three-object subset, the complete hybrid, roller-only (JAM OFF), and jamming-only (ROLLER OFF) conditions achieved 80/81, 54/81, and 24/81 successful trials, respectively. The spatial outcomes of the roller-only and jamming-only conditions are shown in
Fig.~\ref{fig:15}.

%===============================================================================

\subsection{Orientation Offset Robustness Results}
\begin{figure*}[thpb]
  \centering\centering
      \includegraphics[width=\linewidth]{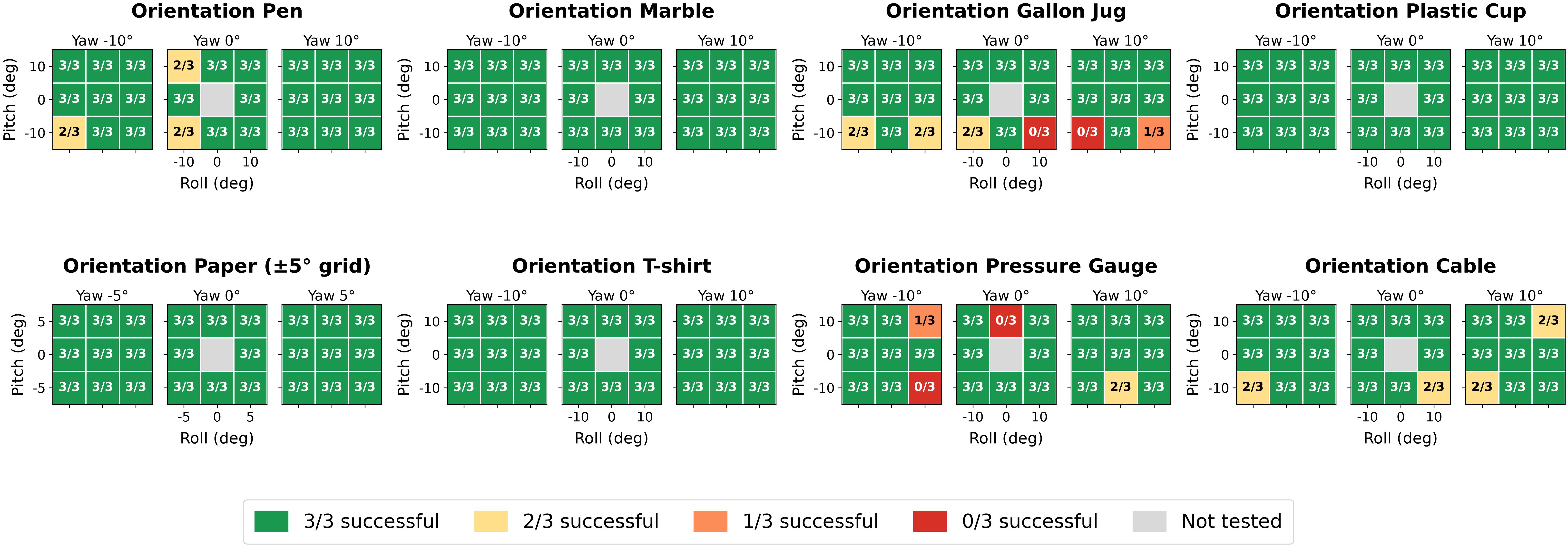}
  \caption{Orientation-offset test results across the roll, pitch, and yaw conditions, with three trials per condition. Each cell represents one orientation-offset condition repeated three times, and the number and colour in each cell indicate the number of successful trials out of three. The $3 \times 3 \times 3$ roll--pitch--yaw grid excludes the all-zero condition, shown in grey as not tested. For the paper sheet, the orientation-offset magnitude was reduced to $\pm 5^{\circ}$ to avoid gripper--table collision.}
  \label{fig:16}
\end{figure*}

Fig.~\ref{fig:16} reports the repeated outcomes across the 26 non-zero roll-pitch-yaw conditions. Each condition was repeated three times, resulting in 78 orientation trials per object. Across all eight objects, the gripper achieved 597 successes in 624 trials, corresponding to an overall orientation-offset success rate of 95.7\%. The marble, plastic cup, T-shirt, and paper sheet showed the highest success rates across their respective tested orientation ranges. The gallon jug and pressure gauge were the most challenging rigid objects. The pen and cable showed intermediate performance, with a smaller number of failures.

%%%%%%%%%%%%%%%%%%%%%%%%%%%%%%%%%%%%%%%%%%%%%%%%%%%%%%%%%%%%%%%%%%%%%%%%%%%%%%%%

\section{Discussion}

\subsection{Interpretation of the Planar-Offset Results}
The repeated planar-offset results indicate that the gripper can recover from the tested planar offsets before jamming is applied. In many successful trials, inward roller motion drew the object closer to the centre region of the spherical membranes, therefore increasing the contact area available before stiffening. The gallon jug accounted for the only failure in the planar-offset protocol. This result suggests that the combination of object mass, off-centre loading, and non-uniform rolling can reduce grasp stability. Because each roller is driven from one end, heavy loading or membrane deformation can cause the driven end to rotate first. This may create a small rotational lag along the roller. Because only one failure occurred in the planar-offset evaluation, further testing under asymmetric loading is needed to determine whether this is a repeatable limitation.

%===============================================================================

\subsection{Interpretation of the Ablation Results}
The ablation results show that rolling and jamming contribute differently depending on the object. The gallon jug performed better with jamming only (ROLLER OFF), whereas the pen and cable performed better with rolling only (JAM OFF).

For the gallon jug, initial contact was usually sufficient because of its large size. The main challenge was retaining the heavy object during lifting and transport. Jamming improved contact stiffness and reduced slip, while the roller-only condition often acquired the jug but failed to hold it. However, the complete hybrid sequence still provided an advantage because rolling improved the object position and membrane contact before jamming. The pen and cable presented the opposite challenge. Their thin and low-profile shapes made initial acquisition difficult without roller intake. In the jamming-only condition, vacuum often stiffened a shallow or incomplete contact rather than drawing the object further into the grasp. Rolling was therefore more important for gathering these objects from partial contact.

Overall, rolling mainly supports object acquisition and contact formation, while jamming improves post-capture retention. The complete hybrid sequence combines these functions by first developing contact through rolling and then stabilising the grasp through jamming.

%===============================================================================

\subsection{Interpretation of the Orientation Offset Results}
The orientation results suggest that the gripper is more sensitive to three-dimensional contact conditions than to planar position offsets. Under orientation offsets, the object surface contacting each roller changes, making it more difficult for rolling to improve the initial grasp. In many failed trials, contact was one-sided or too shallow for the object to be drawn inward before jamming. Because the controller cannot confirm contact with both rollers or adapt the intake duration, jamming may begin while contact is still limited or uneven.

This was most evident for the pressure gauge and gallon jug. Their rigid and asymmetric shapes produced large changes in contact as the approach orientation varied. Under difficult offsets, the membranes sometimes deformed around the initial contact region without moving the object into a better position. Jamming then preserved this poor contact, leaving gaps between the object and membrane and reducing grasp stability.

The paper sheet achieved reliable performance within the reduced $\pm5^{\circ}$ orientation range. This result is not directly comparable with the $\pm10^{\circ}$ range used for the other objects. The smaller range was required to avoid gripper–table collision during the low-height approach. The remaining limitation for the paper sheet is therefore the restricted collision-free approach range rather than a high failure rate within the tested range.

Objects such as the marble, plastic cup, and T-shirt remained easier because their geometry allowed more consistent contact across the tested orientations. The pen and cable were intermediate cases. The few failures of the pen and cable mainly occurred during acquisition at larger combined angles. Their slender or elongated shapes often produced contact on only one side, limiting roller intake before jamming.

%%%%%%%%%%%%%%%%%%%%%%%%%%%%%%%%%%%%%%%%%%%%%%%%%%%%%%%%%%%%%%%%%%%%%%%%%%%%%%%%

\section{Limitations and Future Work}
The present study provides an initial prototype-level evaluation of the proposed mechanism, but several limitations remain in the experimental scope, hardware implementation, and control strategy.

%===============================================================================

\subsection{Experimental Limitations}
Each position-offset and orientation-offset condition was repeated three times. This provides an initial measure of repeatability, but the results and confidence intervals mainly describe the tested conditions and should not be broadly generalised to unseen objects. 

The object set covers several contact and deformation conditions but is not a standardised benchmark. Object mass, geometry, material, and surface friction were also not varied independently. Furthermore, all experiments used a clear tabletop and a top-down approach, without clutter, constrained approaches, or an external baseline gripper.

%===============================================================================

\subsection{Mechanical Limitations}
The current prototype is relatively bulky and heavy, which restricts the available working space around the object. The single-actuator mechanism provides simple, synchronous, and self-locking finger motion. However, the fingers cannot adapt independently to asymmetric or one-sided contact. This can reduce balanced contact formation, particularly for thin objects such as paper. The results also suggest possible non-uniform roller motion along the roller length under high deformation or off-centre loading. Consequently, the current design is less suitable for large or heavy objects that require strong and symmetric power-grasp contact.

%===============================================================================

\subsection{Control and Sensing Limitations}
The controller uses motor-current feedback during \texttt{CLOSE} and \texttt{HOLD}, whereas INTAKE and JAM use fixed durations. Motor current provides only an indirect measure of the overall closing load. It cannot identify which roller is in contact, confirm contact with both rollers, or determine grasp stability. The intake duration, grasp height, and clamping settings are calibrated for each object. This supports consistent evaluation but limits operation with unseen objects. The present system should accordingly be regarded as a mechanism-validation platform rather than a fully autonomous grasping system.

%===============================================================================

\subsection{Modelling Limitations}
The geometric analysis represents the fingers as rigid links and the spherical membrane rollers as undeformed circular envelopes. It does not model membrane indentation, granular redistribution, vacuum-induced stiffness changes, friction, contact pressure, or three-dimensional contact. The derived relationships therefore provide nominal geometric references rather than experimentally validated grasping limits.

%===============================================================================

\subsection{Future Work}
Future work will evaluate the gripper using standardised object sets, controlled variations in object properties, and cluttered or obstacle-constrained environments. Hardware development will focus on reducing the size and mass of the gripper, improving roller-motion uniformity, and introducing adaptive or independently compliant finger closure. Tactile and vacuum-pressure sensing will also be integrated to support event-driven state transitions and reduce object-specific calibration.

%%%%%%%%%%%%%%%%%%%%%%%%%%%%%%%%%%%%%%%%%%%%%%%%%%%%%%%%%%%%%%%%%%%%%%%%%%%%%%%%

\section{Conclusion}
This paper presented a hybrid roller-jamming gripper that combines active rolling intake with granular-jamming stiffening. Across 840 repeated planar-offset and orientation-offset trials on eight objects, the gripper achieved 812 successful grasps. In the corresponding three-object ablation subset, the hybrid, roller-only, and jamming-only conditions achieved 80/81, 54/81, and 24/81 successes, respectively. These results support the complementary roles of rolling in improving
acquisition after partial or off-centre contact and jamming in improving post-capture retention.

The present prototype remains limited by object-specific calibration, partly time-driven control, a controlled tabletop environment, and idealised geometric modelling. Nevertheless, the repeated trials provide mechanism-level evidence for combining active intake with post-contact stiffening. This approach is promising for grasping under position and orientation offsets.

% =========================================================
% ACKNOWLEDGMENT
% =========================================================
% \section*{Acknowledgment}

% Add acknowledgments here if applicable.

% =========================================================
% APPENDIX (OPTIONAL)
% =========================================================
% \appendices
% \section{Additional Results}
% Add supplementary material here if needed.

% =========================================================
% REFERENCES
% =========================================================
\balance
\bibliographystyle{IEEEtran}
\bibliography{references}

@article{de2019grasping,
  title={Grasping objects from the floor in assistive robotics: Real world implications and lessons learned},
  author={De La Puente, Paloma and Fischinger, David and Bajones, Markus and Wolf, Daniel and Vincze, Markus},
  journal={IEEE Access},
  volume={7},
  pages={123725--123735},
  year={2019},
  publisher={IEEE}
}

@article{hsiao2011robust,
  title={Robust grasping under object pose uncertainty},
  author={Hsiao, Kaijen and Kaelbling, Leslie Pack and Lozano-P{\'e}rez, Tom{\'a}s},
  journal={Autonomous Robots},
  volume={31},
  number={2},
  pages={253--268},
  year={2011},
  publisher={Springer}
}

@article{deimel2016novel,
  title={A novel type of compliant and underactuated robotic hand for dexterous grasping},
  author={Deimel, Raphael and Brock, Oliver},
  journal={The International Journal of Robotics Research},
  volume={35},
  number={1-3},
  pages={161--185},
  year={2016},
  publisher={SAGE Publications Sage UK: London, England}
}

@article{chappell2023hydra,
  title={The hydra hand: A mode-switching underactuated gripper with precision and power grasping modes},
  author={Chappell, Digby and Bello, Fernando and Kormushev, Petar and Rojas, Nicolas},
  journal={IEEE Robotics and Automation Letters},
  volume={8},
  number={11},
  pages={7599--7606},
  year={2023},
  publisher={IEEE}
}

@inproceedings{tran2025hybrid,
  title={Hybrid gripper with passive pneumatic soft joints for grasping deformable thin objects},
  author={Tran, Ngoc-Duy and Ly, Hoang-Hiep and Nguyen, Xuan-Thuan and Mac, Thi-Thoa and Nguyen, Anh and Ta, Tung D},
  booktitle={2025 IEEE International Conference on Robotics and Automation (ICRA)},
  pages={7858--7864},
  year={2025},
  organization={IEEE}
}

@article{amend2012positive,
  title={A positive pressure universal gripper based on the jamming of granular material},
  author={Amend, John R and Brown, Eric and Rodenberg, Nicholas and Jaeger, Heinrich M and Lipson, Hod},
  journal={IEEE Transactions on Robotics},
  volume={28},
  number={2},
  pages={341--350},
  year={2012},
  publisher={IEEE}
}

@article{zeng2023high,
  title={A high performance pneumatically actuated soft gripper based on layer jamming},
  author={Zeng, Xianpai and Su, Hai-Jun},
  journal={Journal of Mechanisms and Robotics},
  volume={15},
  number={1},
  pages={014501},
  year={2023},
  publisher={American Society of Mechanical Engineers}
}

@article{piskarev2023soft,
  title={A soft gripper with granular jamming and electroadhesive properties},
  author={Piskarev, Yegor and Devincenti, Antoine and Ramachandran, Vivek and Bourban, Pierre-Etienne and Dickey, Michael D and Shintake, Jun and Floreano, Dario},
  journal={Advanced Intelligent Systems},
  volume={5},
  number={6},
  pages={2200409},
  year={2023},
  publisher={Wiley Online Library}
}

@inproceedings{keller2024phase,
  title={A Phase-Change Emulsion Jamming Gripper for Manipulation of Micro-Scale Textured Surfaces},
  author={Keller, Alex and Yue, Tianqi and Qi, Qiukai and Conn, Andrew T and Rossiter, Jonathan},
  booktitle={2024 IEEE International Conference on Robotics and Automation (ICRA)},
  pages={706--712},
  year={2024},
  organization={IEEE}
}

@article{hou2019design,
  title={Design and experiment of a universal two-fingered hand with soft fingertips based on jamming effect},
  author={Hou, Taogang and Yang, Xingbang and Aiyama, Yasumichi and Liu, Kaiqi and Wang, Zeyu and Wang, Tianmiao and Liang, Jianhong and Fan, Yubo},
  journal={Mechanism and Machine Theory},
  volume={133},
  pages={706--719},
  year={2019},
  publisher={Elsevier}
}

@article{amend2017jamhand,
  title={The {JamHand}: dexterous manipulation with minimal actuation},
  author={Amend, John and Lipson, Hod},
  journal={Soft Robotics},
  volume={4},
  number={1},
  pages={70--80},
  year={2017},
  publisher={SAGE Publications Sage CA: Los Angeles, CA}
}

@article{badilla2024hybgrip,
  title={{HybGrip}: A synergistic hybrid gripper for enhanced robotic surgical instrument grasping},
  author={Badilla-Sol{\'o}rzano, Jorge and Ihler, Sontje and Seel, Thomas},
  journal={International Journal of Computer Assisted Radiology and Surgery},
  volume={19},
  number={12},
  pages={2363--2370},
  year={2024},
  publisher={Springer}
}

@article{kim2024development,
  title={Development of a hybrid gripper with jamming module using small-scale pneumatic pump},
  author={Kim, Myeongjin and Yoon, Jingon and Kim, Donghyun and Yun, Dongwon},
  journal={International Journal of Control, Automation and Systems},
  volume={22},
  number={11},
  pages={3341--3351},
  year={2024},
  publisher={Springer}
}

@inproceedings{mizushima2018multi,
  title={Multi-fingered robotic hand based on hybrid mechanism of tendon-driven and jamming transition},
  author={Mizushima, Kaori and Oku, Takumi and Suzuki, Yosuke and Tsuji, Tokuo and Watanabe, Tetsuyou},
  booktitle={2018 IEEE International Conference on Soft Robotics (RoboSoft)},
  pages={376--381},
  year={2018},
  organization={IEEE}
}

@article{qin2025rolling,
  title={From rolling-in to enveloping: An active gripper for efficient handling of spherical fruits in agriculture},
  author={Qin, Huanhuan and Qiu, Zicheng and Zhu, Haoran and Diao, Rui and Che, Fengwei and Gu, Xingjian and Luo, Yan and Zhang, Baohua and Lu, Mingzhou},
  journal={Computers and Electronics in Agriculture},
  volume={239},
  pages={110995},
  year={2025},
  publisher={Elsevier}
}

@article{xie2023hand,
  title={In-hand manipulation with a simple belted parallel-jaw gripper},
  author={Xie, Gregory and Holladay, Rachel and Chin, Lillian and Rus, Daniela},
  journal={IEEE Robotics and Automation Letters},
  volume={9},
  number={2},
  pages={1334--1341},
  year={2024},
  publisher={IEEE}
}

@inproceedings{yuan2020design,
  title={Design of a roller-based dexterous hand for object grasping and within-hand manipulation},
  author={Yuan, Shenli and Epps, Austin D and Nowak, Jerome B and Salisbury, J Kenneth},
  booktitle={2020 IEEE International Conference on Robotics and Automation (ICRA)},
  pages={8870--8876},
  year={2020},
  organization={IEEE}
}

@inproceedings{yuan2020designV2,
  title={Design and control of roller grasper v2 for in-hand manipulation},
  author={Yuan, Shenli and Shao, Lin and Yako, Connor L and Gruebele, Alex and Salisbury, J Kenneth},
  booktitle={2020 IEEE/RSJ International Conference on Intelligent Robots and Systems (IROS)},
  pages={9151--9158},
  year={2020},
  organization={IEEE}
}

@article{yuan2024designV3,
  title={Design and control of roller grasper v3 for in-hand manipulation},
  author={Yuan, Shenli and Shao, Lin and Feng, Yunhai and Sun, Jiatong and Xue, Teng and Yako, Connor L and Bohg, Jeannette and Salisbury, J Kenneth},
  journal={IEEE Transactions on Robotics},
  volume={40},
  pages={4222--4234},
  year={2024},
  publisher={IEEE}
}

@article{li2026dexterous,
  title={A Dexterous Hand for Omnidirectional In-Hand Manipulation: Design, Analysis and Experimental Validation.},
  author={Li, Huaiyong and Ye, Changlong and Jia, Rongdian and Yu, Suyang and Tao, Guanghong},
  journal={Biomimetics (Basel, Switzerland)},
  volume={11},
  number={3},
  pages={167},
  year={2026}
}

@article{falco2020benchmarking,
  title={Benchmarking protocols for evaluating grasp strength, grasp cycle time, finger strength, and finger repeatability of robot end-effectors},
  author={Falco, Joe and Hemphill, Daniel and Kimble, Kenny and Messina, Elena and Norton, Adam and Ropelato, Rafael and Yanco, Holly},
  journal={IEEE Robotics and Automation Letters},
  volume={5},
  number={2},
  pages={644--651},
  year={2020},
  publisher={IEEE}
}

@inproceedings{matheus2010benchmarking,
  title={Benchmarking grasping and manipulation: Properties of the objects of daily living},
  author={Matheus, Kayla and Dollar, Aaron M},
  booktitle={2010 IEEE/RSJ International Conference on Intelligent Robots and Systems},
  pages={5020--5027},
  year={2010},
  organization={IEEE}
}

@inproceedings{calli2015ycb,
  title={The {YCB} object and model set: Towards common benchmarks for manipulation research},
  author={Calli, Berk and Singh, Arjun and Walsman, Aaron and Srinivasa, Siddhartha and Abbeel, Pieter and Dollar, Aaron M},
  booktitle={2015 International Conference on Advanced Robotics (ICAR)},
  pages={510--517},
  year={2015},
  organization={IEEE}
}

\end{document}